%% file: main.tex
\documentclass[10pt, journal,compsoc]{IEEEtran}
\usepackage[T1]{fontenc}  
\usepackage{amsmath,amsfonts}
\usepackage{algorithm}
\usepackage{algorithmicx}
\usepackage{algpseudocode}
\usepackage{multirow,array}
\usepackage[font=normalsize]{caption}
\usepackage{subcaption}
\usepackage{textcomp}
\usepackage{stfloats}
\usepackage{url}
\usepackage{verbatim}
\usepackage{graphicx}
\usepackage{cite}
\usepackage{booktabs}
\usepackage{enumitem}
\usepackage[dvipsnames]{xcolor}
\usepackage[most]{tcolorbox}
\usepackage{float}
\usepackage{tikz}
\usepackage{forest}
\usepackage{fancyhdr} 
\usetikzlibrary{trees,positioning,shapes,mindmap,shadows,arrows.meta}
\renewcommand{\theparagraph}{\arabic{paragraph}}
\usepackage{titlesec}
\titleformat{\paragraph}
[runin]
{\bfseries}
{\theparagraph{})}  % 这里显示编号
{1em}
{}
\titlespacing{\paragraph}
{0pt}{0pt}{1em}
\begin{document}
% When LLMs Form Teams: The Emergence of Collaborative Affective Intelligence
% The Emergence of Collaborative Affective Computing: A Survey
\title{When LLMs Team Up: The Emergence of Collaborative Affective Computing}
\author{Wenna Lai, Haoran Xie, Guandong Xu, Qing Li, S. Joe Qin
\thanks{The research described in this article has been supported by a grant from the Research Grants Council of the Hong Kong Special Administrative Region, China (R1015-23), and the Faculty Research Grant (SDS24A8) and the Direct Grant (DR25E8) of Lingnan University, Hong Kong. \emph{(Corresponding author: Haoran Xie.)}}
\IEEEcompsocitemizethanks{
    \IEEEcompsocthanksitem Wenna Lai is with the Department of Computing, Hong Kong Polytechnic University, Hong Kong SAR (email: winnelai05@gmail.com).
    \IEEEcompsocthanksitem Haoran Xie is with the School of Data Science, Lingnan University, Hong Kong SAR (email: hrxie@ln.edu.hk).
    \IEEEcompsocthanksitem Guandong Xu is with the School of Computer Science and the Data Science Institute, University of Technology Sydney, Sydney, NSW 2007, and also the Education University of Hong Kong, Hong Kong SAR (e-mail: gdxu@eduhk.hk).
    \IEEEcompsocthanksitem Qing Li is with the Department of Computing, Hong Kong Polytechnic University, Hong Kong SAR (e-mail: qing-prof.li@polyu.edu.hk).
    \IEEEcompsocthanksitem S. Joe Qin is with the School of Data Science, Lingnan University, Hong Kong SAR (email: joeqin@ln.edu.hk).
}
}



\IEEEtitleabstractindextext{%
\begin{abstract}
Affective Computing (AC) is essential in bridging the gap between human emotional experiences and machine understanding. Traditionally, AC tasks in natural language processing (NLP) have been approached through pipeline architectures, which often suffer from structure rigidity that leads to inefficiencies and limited adaptability. The advent of Large Language Models (LLMs) has revolutionized this field by offering a unified approach to affective understanding and generation tasks, enhancing the potential for dynamic, real-time interactions. However, LLMs face cognitive limitations in affective reasoning, such as misinterpreting cultural nuances or contextual emotions, and hallucination problems in decision-making. To address these challenges, recent research advocates for LLM-based collaboration systems that emphasize interactions among specialized models and LLMs, mimicking human-like affective intelligence through the synergy of emotional and rational thinking that aligns with \emph{Dual Process Theory} in psychology. This survey aims to provide a comprehensive overview of LLM-based collaboration systems in AC, exploring from structured collaborations to autonomous collaborations. Specifically, it includes: (1) A systematic review of existing methods, focusing on collaboration strategies, mechanisms, key functions, and applications; (2) Experimental comparisons of collaboration strategies across representative tasks in affective understanding and generation; (3) An analysis highlighting the potential of these systems to enhance robustness and adaptability in complex affective reasoning; (4) A discussion of key challenges and future research directions to further advance the field. This work is the first to systematically explore collaborative intelligence with LLMs in AC, paving the way for more powerful applications that approach human-like social intelligence.
\end{abstract}

}

\maketitle
\IEEEpeerreviewmaketitle

\input{src/introduction}

\input{src/Background}
\input{src/collaboration_framework}

\input{src/experiment}
\input{src/discussion}

\input{src/conclusion}


\bibliographystyle{IEEEtran}
\bibliography{IEEEabrv, main}
%

% \vspace{100pt}

\begin{IEEEbiography}[{\includegraphics[width=1in, height=1.25in,clip,keepaspectratio]{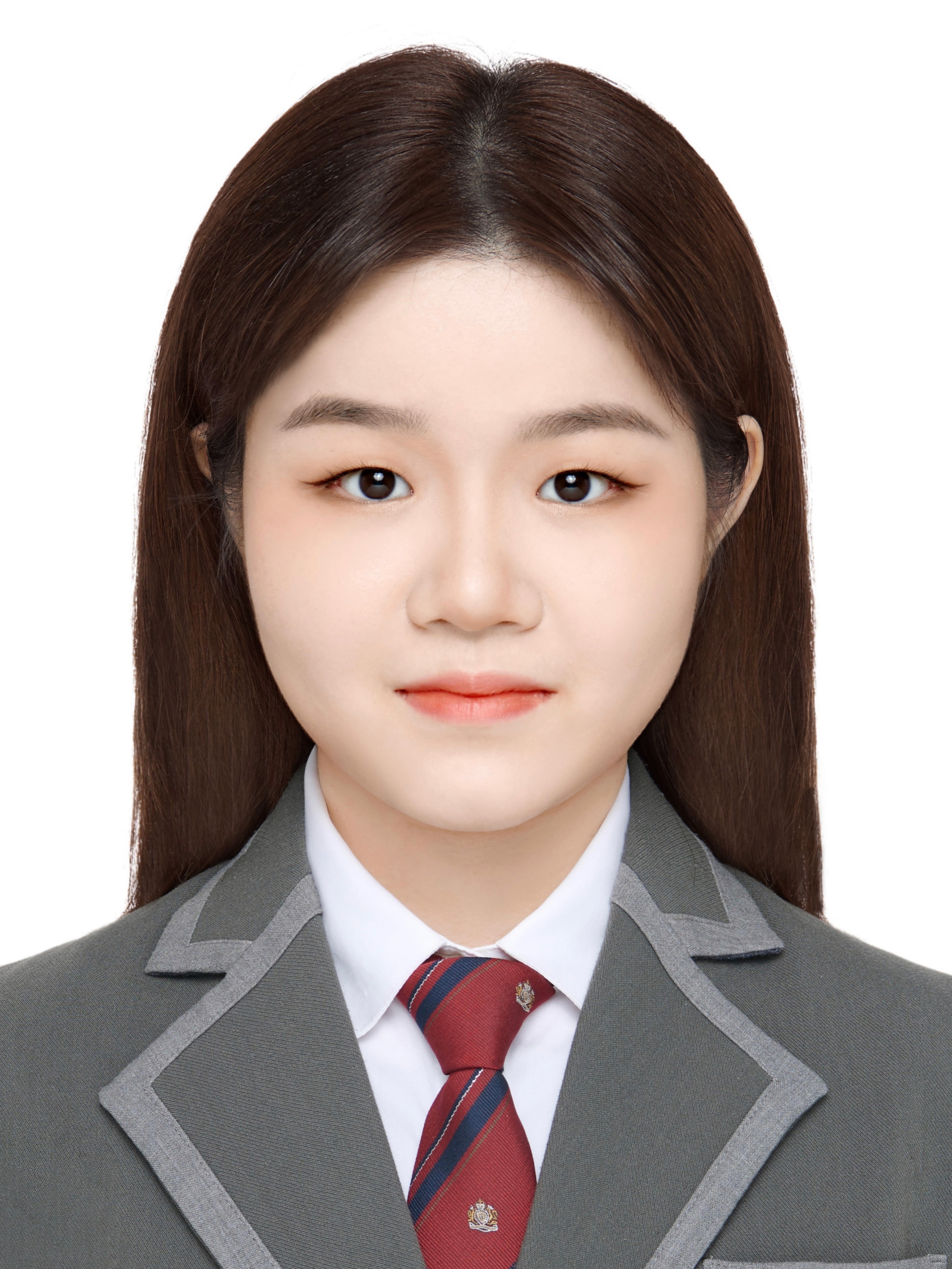}}]{Wenna Lai} (Student Member, IEEE)
is currently a Ph.D. student at the Department of Computing, Hong Kong Polytechnic University, under the supervision of Prof. Qing Li. She has been working closely with Prof. Guandong Xu at the School of Computer Science, University of Technology Sydney, and Prof. Haoran Xie at the School of Data Science, Lingnan University, Hong Kong. Before that, she received her Master's degree in the Department of Electrical and Computer Engineering from the National University of Singapore. Her research interests include Affective Computing and NLP for Social Good.
\end{IEEEbiography}
\vspace{-5pt}
% \vspace{-1em}
\begin{IEEEbiography}[{\includegraphics[width=1in,height=1.25in,clip,keepaspectratio]{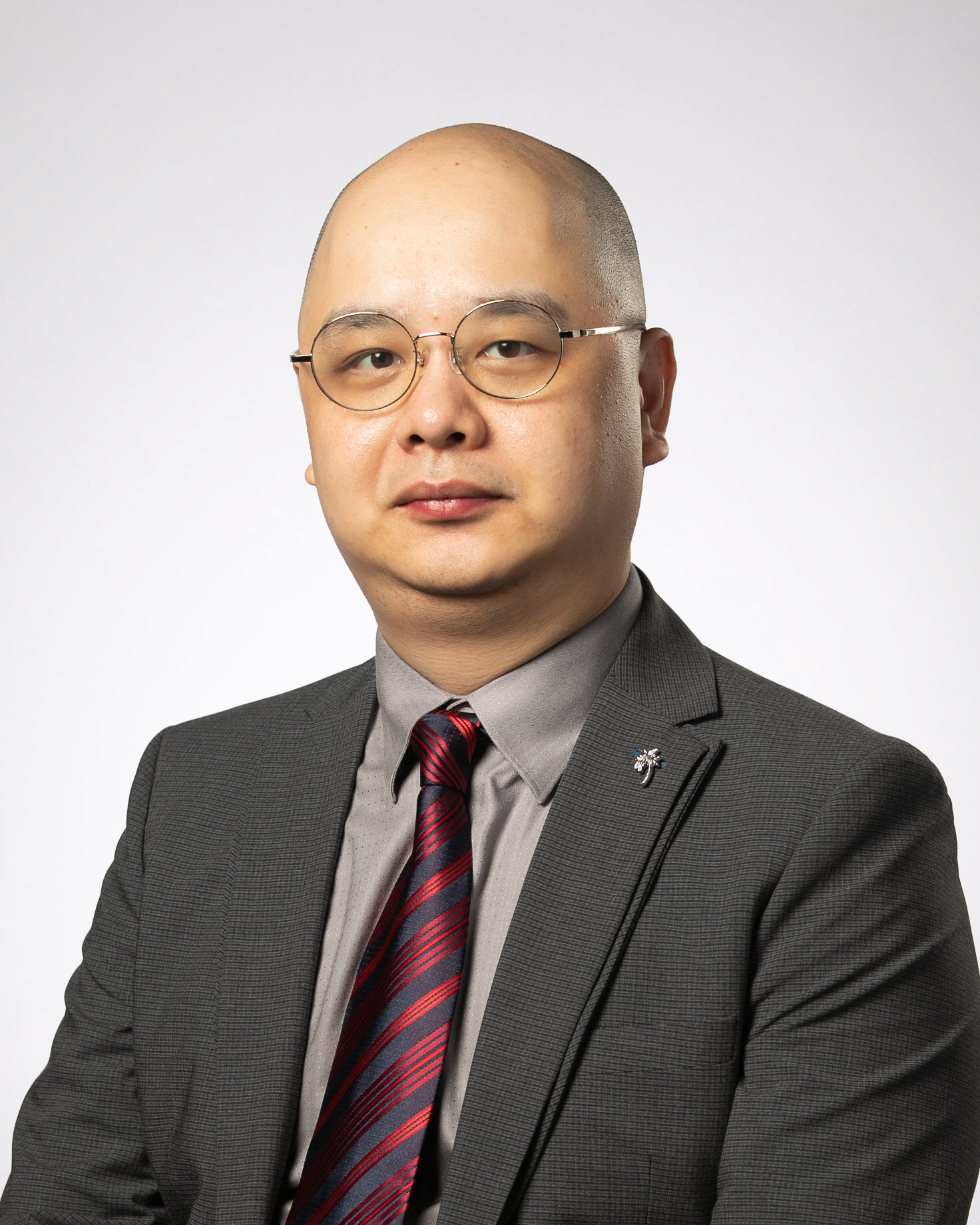}}]{Haoran Xie} (Senior Member, IEEE)
received a Ph.D. degree in Computer Science from City University of Hong Kong and an Ed.D degree in Digital Learning from the University of Bristol. He is currently a Professor and the Person-in-Charge at the Division of Artificial Intelligence, Director of LEO Dr David P. Chan Institute of Data Science, and Associate Dean of the School of Data Science, Lingnan University, Hong Kong. His research interests include natural language processing, large language models, language learning, and AI in education. He has published 450 research publications, including 260 journal articles. He is the Editor-in-Chief of Natural Language Processing Journal, Computers \& Education: Artificial Intelligence, and Computers \& Education: X Reality. He has been selected as the World's Top 2\% Scientists by Stanford University.
\end{IEEEbiography}
\vspace{-5pt}
% \vspace{-1em}
\begin{IEEEbiography}[{\includegraphics[width=1in,height=1.25in,clip,keepaspectratio]{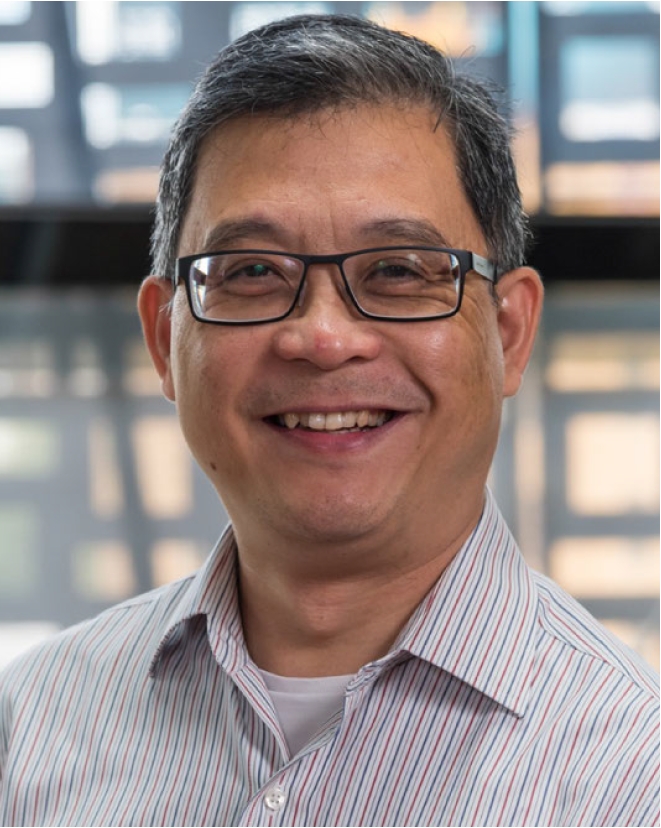}}]{Guandong Xu} (Member, IEEE) received the Ph.D. degree in computer science from Victoria University, Melbourne, VIC, Australia, in 2009. He is currently a Professor and a Program Leader at the School of Computer Science and Data Science Institute, University of Technology Sydney, Sydney, NSW, Australia. His research interests include data science, data analytics, recommender systems, web mining, user modeling, NLP, social network analysis, and social media mining.
\end{IEEEbiography}
\vspace{-5pt}
% \vspace{-1em}
\begin{IEEEbiography}[{\includegraphics[width=1in,height=1.25in,clip,keepaspectratio]{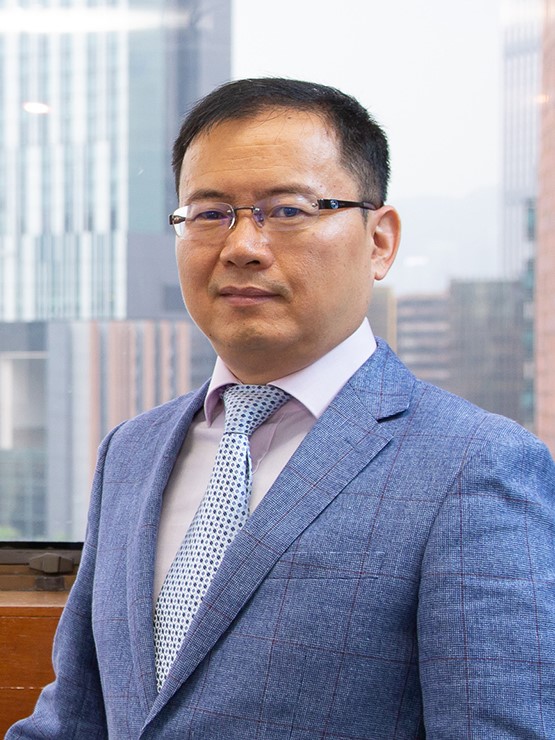}}]{Qing Li} (Fellow, IEEE) received the B.Eng. degree in Computer Science from Hunan Univeristy, Hunan, China, in 1982, and the M.S. and Ph.D. degrees in Computer Science from the University of Southern California, LA, California, USA, in 1985 and 1988, respectively. Qing Li is a Chair Professor and Head at the Department of Computing, The Hong Kong Polytechnic University. His research focuses on data science, web mining, and artificial intelligence. He is a Fellow of IET, a Fellow of IEEE, a member of ACM SIGMOD and IEEE Technical Committee on Data Engineering. He is the chairperson of the Hong Kong Web Society, and is a steering committee member of DASFAA, ICWL, and WISE Society.
\end{IEEEbiography}
\vspace{-5pt}
% \vspace{-1em}
\begin{IEEEbiography}[{\includegraphics[width=1in,height=1.25in,clip,keepaspectratio]
{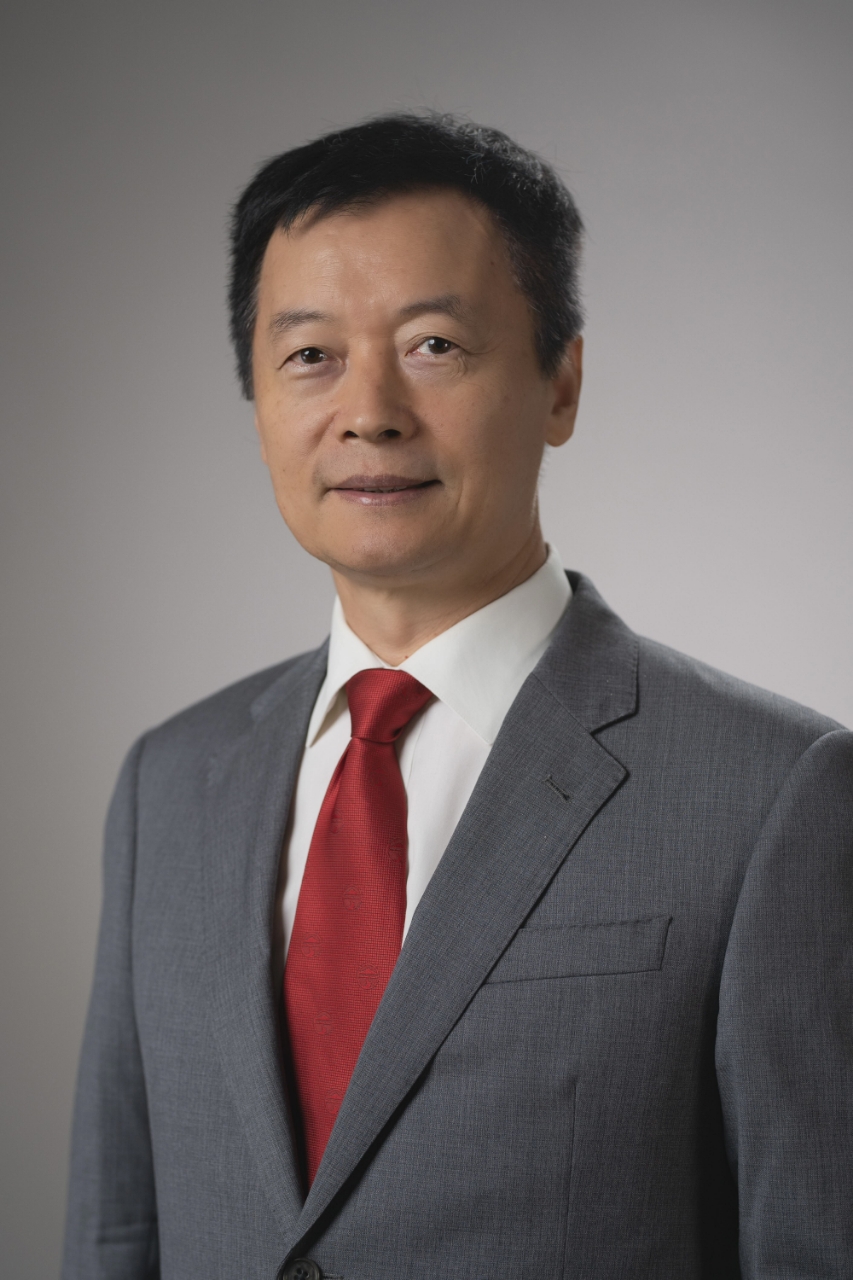}}]
{S. Joe Qin} (Fellow, IEEE) received the B.S. and M.S. degrees in automatic control from Tsinghua University, Beijing, China, in 1984 and 1987, respectively, and the Ph.D. degree in chemical engineering from the University of Maryland, College Park, MD, USA, in 1992. He is currently the Wai Kee Kau Chair Professor and President of Lingnan University, Hong Kong. His research interests include data science and analytics, machine learning, process monitoring, model predictive control, system identification, smart manufacturing, smart cities, and predictive maintenance. Prof. Qin is a Fellow of the U.S. National Academy of Inventors, IFAC, and AIChE. He was the recipient of the 2022 CAST Computing Award by AIChE, 2022 IEEE CSS Transition to Practice Award, U.S. NSF CAREER Award, and NSF-China Outstanding Young Investigator Award. His h-indices for Web of Science, SCOPUS, and Google Scholar are 66, 73, and 89, respectively.
\end{IEEEbiography}
\vfill

\end{document}

%% file: src/introduction.tex
\section{Introduction}
\thispagestyle{plain}
Affective Computing (AC), first conceptualized by Rosalind Picard in 1997 \cite{picard}, emerged as a discipline to bridge the gap between human emotional experience and machine understanding. Early works defined AC as ``computing that relates to, arises from, or deliberately influences emotions'' \cite{pami/PicardVH01}, emphasizing its role in enabling machines to perceive, interpret, and simulate human affects, including emotions, moods, and feelings \cite{Russellaffect, affect}. The academic community increasingly recognizes that emotions are inherently social phenomena, as they are typically elicited, expressed, regulated, perceived, interpreted, and responded to within social contexts \cite{vanKleef2016EditorialTS}. To achieve human-like intelligence, affective computing is considered an essential component in understanding and interpreting these complex emotional processes \cite{ACandSA, WangSTLYLGSGZZ22, access/AfzalKPL24, acii/TaoT05}, which transformed human-computer interaction from logic-driven dialogue to affect-sensitive communication \cite{McColl2015ASO, Scheutz}.

\begin{figure*}[h!]
\centering
\includegraphics[width=\textwidth]{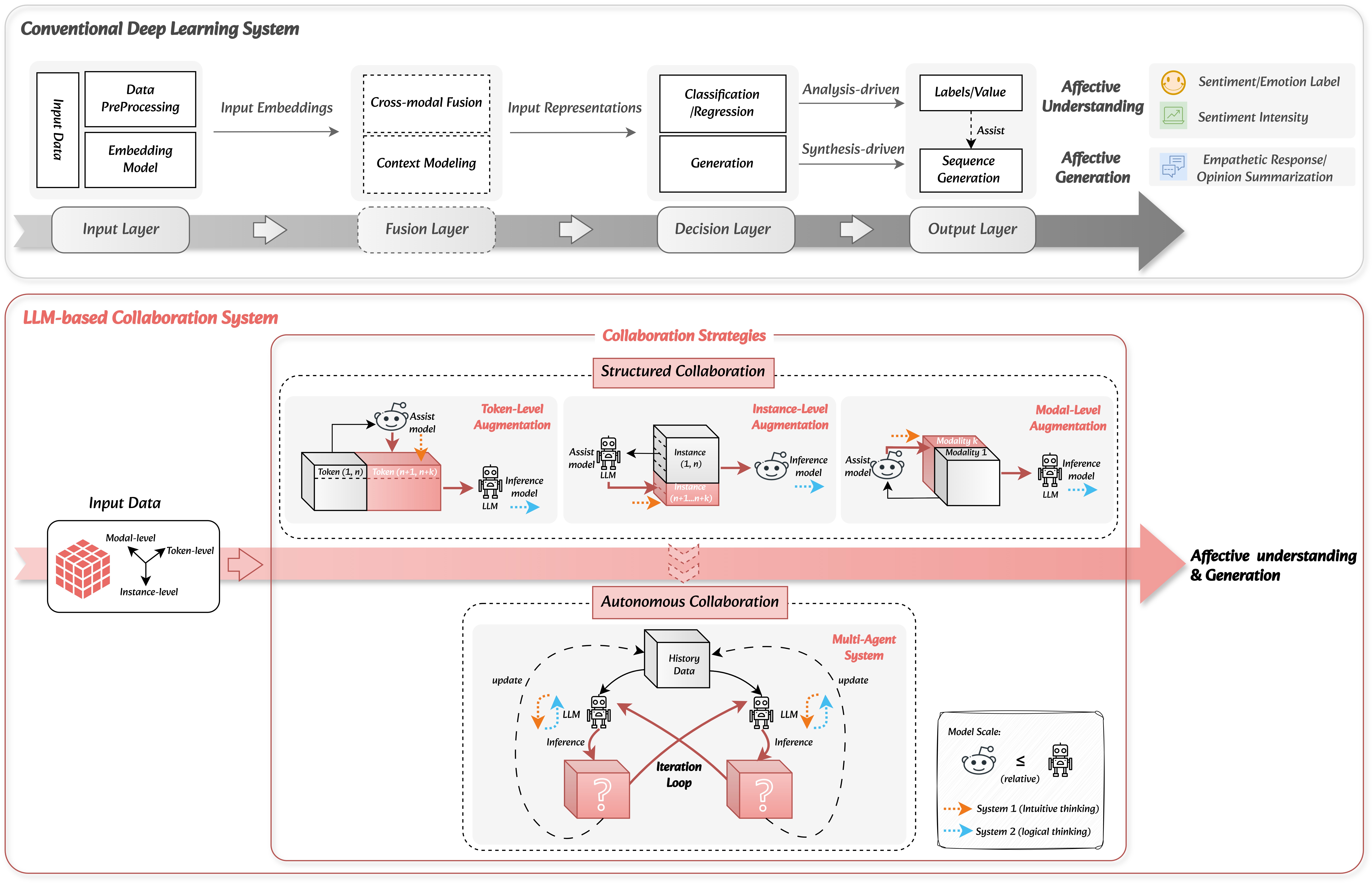}
\caption{An overview of the LLM-based collaboration system compared to the conventional deep learning system in affective computing. The strategies for LLM-based collaboration are aligned with \emph{Dual Process Theory} and can be systematically classified, progressing from structured to autonomous collaboration, according to model interaction patterns.}
\label{fig:framework}
\end{figure*}

The AC tasks can be generally categorized into two dominant tasks in natural language processing (NLP), which are affective understanding (AU) and affective generation (AG) \cite{Zhang2024AffectiveCI, gptEmotionalCapabilities, peiIC}. AU tasks, which are \emph{analysis-driven}, typically involve predicting class labels or scalar values that indicate affective information, such as sentiment analysis \cite{rvisa, Zhang2023SentimentAI}, and emotional intensity detection \cite{taffco/AminMCS24, AlonsoCMT15}. In contrast, AG tasks, which are \emph{synthesis-driven}, aim to generate sequences with emotional nuances, such as empathetic response generation \cite{aaai/SabourZH22, SahaIJCNN}.  Over the past decades, conventional deep learning systems have approached affective computing tasks through a pipeline architecture, as illustrated in Figure \ref{fig:framework}. These systems handle inputs through sequential stages: (1) \emph{input layer} processes inputs through embedding models to generate embedding vectors; (2) \emph{fusion layer} employs context modeling or cross-modal fusion techniques on embeddings to develop input representations conducive to subsequent task learning; (3) \emph{decision layer} applies distinct loss functions and output architectures according to affective task types; (4) \emph{output layer} predicts discrete labels or generates affect-aware sequences accordingly. In such systems, the affective computing tasks can be addressed end-to-end through the appropriate design of deep neural networks \cite{pami/PicardVH01, SequentialSA, EmpMFF, ijcai/00080XXSL22}. Some research has also explored enhancing AG tasks with fine-grained information from AU tasks, such as using emotional labels to condition dialogue models \cite{acl/RashkinSLB19, wang-etal-2022-empathetic}. However, such pipeline systems exhibit a clear division of roles across neural layers and hierarchical decoupling to incrementally achieve task objectives, with each functional layer completing its task independently. This siloed design can lead to parameter isolation and error propagation, where layers cannot share gradients or adapt holistically during training, reducing efficiency and interpretability. Moreover, AU and AG tasks are treated separately in terms of training objectives and paradigms, limiting dynamic adaptation to contextual emotional shifts in real-time interactions.

With the advent of Large Language Models (LLMs), there has been a revolutionary change from traditional deep learning to LLMs application in the NLP domain, as LLMs demonstrate remarkable generalization performance across multiple tasks without the stringent requirement for task-specific training \cite{Kojima_Shixiang_Gu_Reid_Matsuo_Iwasawa, Wei2021FinetunedLM, Zhang2023SentimentAI, Zhang2024AffectiveCI}. Unlike conventional deep learning systems, LLMs offer a feasible approach to unify AU and AG tasks as generative tasks \cite{Wang2022UnifiedABSAAU, hu-etal-2022-unimse, legoabsa}. This shift enables LLMs to serve as core models with a unified input semantic space, sharing parameters across diverse tasks. As more advanced LLMs are equipped with self-refinement capabilities when provided reliable external feedback \cite{nips/refine, tacl/KamoiZZHZ24}, they further facilitate end-to-end anthropomorphic interaction characterized by the sequence ``emotion perception → decision → feedback'' as an affective agent \cite{Scheutz, nips/XieCJYLSGBHJE0G24, tran2025multiagentcollaborationmechanismssurvey}, thereby realizing a closed-loop system for affective understanding and generation through dynamic role adjustment.

However, off-the-shelf LLMs often encounter cognitive limitations in affective reasoning, such as conflating contextual and situational emotions or misinterpreting cultural nuances \cite{emnlp/AdilazuardaMLSA24, nips/LiC0S024}. Additionally, LLMs are prone to inherent hallucination problems in their generated responses \cite{Zhang2023SirensSI, Li2024ASO}. To address these challenges, recent research draws on Kahneman’s \emph{Dual Process Theory} \cite{dualprocess}, which suggests that human thought is driven by two systems: \emph{system 1}, an implicit process that is fast, intuitive, and emotional, and \emph{system 2}, an explicit process that is slower, more deliberative and logical. This framework has inspired the development of \textbf{LLM-based collaboration systems} that emphasize collaborative efforts in solving complex reasoning tasks \cite{Schleiger31122024, HeCollaborative, zhang2024collaborativeaisentimentanalysis, tran2025multiagentcollaborationmechanismssurvey, chen2025rolesmallmodelsllm, wang2024comprehensivesurveysmalllanguage}. 

Building on the theoretical foundation, AC inherently requires the simultaneous application of emotional and rational thinking, where \textbf{emotional thinking} intuitively perceives real-time emotions, while \textbf{rational thinking} conducts a deeper analysis of the causes and nuances of these emotions. The \emph{Affective Intelligence Theory} \cite{marcus2000affective} supports this perspective and highlights the inherent complementarity of emotion and reason in human judgment processes. LLM-based collaboration systems leverage this complementarity by synergizing specialized models with LLMs to mimic human-like affective intelligence. For instance, in structured collaboration illustrated in Figure \ref{fig:framework}, an emotion model (\emph{System 1}) might first identify high-arousal emotions and cause words in user messages, then prompt an LLM (\emph{System 2}) to explore potential causes for the emotions before generating a calibrated response \cite{yang2024enhancingerg}. These collaborative approaches \textbf{help mitigate knowledge solidification} in static LLMs while preserving their generative versatility. Additionally, they \textbf{enhance the robustness of complex affective reasoning} through role division, dynamic adaptation, and knowledge sharing. By teaming up more powerful models, LLM-based collaboration systems further \textbf{pave the way for social intelligence} \cite{ASIinTeam} through interactions among smart agents.

Recent efforts have been made to explore LLM-based collaboration, including the development of LLM-based multi-agent systems \cite{Li2024multiagent, tran2025multiagentcollaborationmechanismssurvey, ijcai/GuoCWCPCW024}, as well as interactions with smaller models, which are relatively less resource-intensive compared to their larger counterparts \cite{chen2025rolesmallmodelsllm, wang2024comprehensivesurveysmalllanguage, lee-etal-2024-small}. Both forms of collaboration significantly contribute to the advancement of collaborative AI applications. Instead of limited discussion on a specific form, this survey provides a comprehensive overview, ranging from structured collaboration, including the interaction between LLMs and smaller models, to autonomous collaboration exemplified by LLM-based multi-agent systems. In the domain of affective computing, which is characterized by the collaboration of emotional and logical reasoning, existing reviews primarily focus on the applications and challenges related to specific tasks \cite{access/AfzalKPL24, Cui_Wang_Ho_Cambria_2023, Das2023MultimodalSA, Zhang2022ASO} or the use of LLMs in applications \cite{Zhang2023SentimentAI, Zhang2024AffectiveCI}. To the best of our knowledge, this work is the first to systematically discuss collaborative intelligence with LLMs in affective computing based on the \emph{Dual Process Theory} and \emph{Affective Intelligence Theory}, aiming to facilitate more powerful applications that approach human-like intelligence.

This survey offers a comprehensive overview of LLM-based collaboration systems in affective computing by summarizing existing methods from the perspectives of the collaboration strategies, mechanisms, key functions, and application areas. Specifically, following a brief introduction to the background
knowledge of affective computing tasks and LLMs in Section \ref{background}, we review existing research on building LLM-based collaboration systems for affective computing in terms of structured collaboration and autonomous collaboration in Section \ref{framework}, where structured collaboration encompasses token-level, instance-level, and modal-level augmentation strategies, while autonomous collaboration involves multi-agent systems. Subsequently, we select three representative tasks that cover both affective understanding and generation to conduct experimental comparisons of different collaboration strategies. The results and detailed analyses are presented in Section \ref{experiments}. We provide a comprehensive discussion of the experiments, highlighting key challenges and potential directions for future exploration in Section \ref{discuss}, and draw a conclusion in Section \ref{conclusion}.

%% file: src/Background.tex
\section{Background}
\label{background}
\subsection{Affective Computing}

Affective Computing (AC) is an interdisciplinary field that combines elements from computer science, psychology, and cognitive science to enable machines to recognize, interpret, and simulate human emotions \cite{Russellaffect, affect}. The objective is to develop systems that can understand and respond to human emotions in ways that enhance user interaction and experience across various applications. In the NLP domain, AC can be broadly categorized into two primary tasks: Affective Understanding (AU) and Affective Generation (AG) \cite{Zhang2024AffectiveCI}, which are detailed in the following sections. 

\subsubsection{Affective Understanding}

Affective Understanding involves the recognition and interpretation of emotions from various inputs. Key tasks in AU include sentiment analysis (SA), which involves categorizing emotional expressions as positive, negative, or neutral \cite{Zhang2023SentimentAI, sun2023sentimentanalysisllmnegotiations}. This includes various subtasks such as sentiment classification (SC) that typically focus on sentence-level classification of sentiment categories \cite{Zhang2023SentimentAI, sun2023sentimentanalysisllmnegotiations, ACMse/StigallKANP24}, aspect-based sentiment analysis (ABSA) that analyzes sentiment towards more fine-grained targets or aspect terms in the given text \cite{Gou2023MvPMP, lai2025star, instructabsa, legoabsa, Hoang2019AspectBasedSA, Wang2022UnifiedABSAAU}, emotional intensity detection (EID) that quantifies the intensity of detected emotion \cite{AlonsoCMT15, naacl/KajiwaraCTNN21}, and emotion cause paired extraction (ECPE) that identifies not only emotions but also their potential causes within a given text \cite{taffco/HuZL24, taffco/ChenLLXWW23, coling/GuZMLGTX24}. Other significant tasks have subjective text analysis that involves different aspects of human subjective feeling reflected in the text \cite{BingLiuSentiment, Poria2020BeneathTT}. This includes sarcasm detection (SD), which challenges models to grasp implied sentiments beyond literal meanings \cite{aiopen/MisraA23, aaai/YaoZLQ25, SASD, comsis/ZhouZWWWL25}, and Emotional Recognition in Conversation (ERC), which assesses emotional states throughout dialogues \cite{lei2024instructerc, zhang2024dialoguellm, hu-etal-2022-unimse}. 

Traditional approaches to AU relied heavily on handcrafted features and supervised learning methods, often yielding substantial performance improvements in specific conditions. However, these models can struggle with generalization across diverse contexts, especially when dealing with subtle emotional expressions that vary significantly between cultures \cite{Zhang2024AffectiveCI, picard}. With the rise of Large Language Models (LLMs), substantial advancements have been made in AU tasks. LLMs leverage vast datasets and context-rich representations to perform zero-shot and few-shot learning \cite{Brown2020LanguageMA, Wang_Wei_Schuurmans_Le_Chi_Zhou, Fu_Peng_Sabharwal_Clark_Khot_2022, Zhang_Zhang_Li_Smola_2022}, significantly enhancing the capability to discern and interpret complex emotional cues. While LLMs demonstrate robustness in simpler AU tasks, their performance can wane in more intricate scenarios requiring a deep understanding of context and nuanced emotional expressions \cite{bhaskar-etal-2023-prompted, Zhang2023SentimentAI}.

\subsubsection{Affective Generation}

Affective Generation refers to the creation of content that not only responds to user emotions but is designed to evoke specific emotional reactions \cite{Zhang2024AffectiveCI, gptEmotionalCapabilities, peiIC}. This contains tasks such as empathetic response generation (ERG) \cite{acl/YangRYSCZL24, EmpMFF, cai-etal-2024-empcrl, acl/RashkinSLB19, acl/ZhouZW0H23} and emotional support conversation (ESC) \cite{Peng2022ControlGU, Zheng2022AugESCDA, Liu2021TowardsES}, which are dominant research areas that focus on providing empathetic and emotional assistance in conversation, particularly in sensitive contexts like mental health \cite{chu2024multimodalemo, Deng2023KnowledgeenhancedMD, Cheng2022PALPE}. Other prominent tasks include opinion summarization \cite{Wang2024IterativelyCP, Hosking2024HierarchicalIF, Siledar2024ProductDA}, which involves distilling multiple opinions to highlight the overall sentiment and key points on a particular topic from social media content or reviews.

Traditional generative models struggled with producing emotionally nuanced responses, often leading to output that lacked empathy or relevance to the user's emotional state \cite{cheng-etal-2022-improving}. Generally, AG research progressed more slowly than AU research because of limited model capabilities prior to the advent of LLMs \cite{Zhang2024AffectiveCI}. The introduction of LLMs into affective generation tasks has catalyzed a transformation in how machines produce emotionally intelligent responses. These models have demonstrated the ability to generate text that resonates with user feelings, enhancing user engagement in applications like chatbots, social media, and mental health support systems \cite{Bilquise2022EmotionallyIC, zheng-etal-2024-self}. However, affective generation still faces challenges, such as maintaining consistency in emotional tone and the risk of generating responses that misinterpret user sentiment \cite{acl/KangKKMCYLY24, gptEmotionalCapabilities}.

\subsection{Large Language Models}

Large Language Models (LLMs) represent a significant leap in natural language processing capabilities, driven by advancements in deep learning architectures and massive training datasets. Models such as GPT-3, GPT-4, and LLaMA \cite{touvron2023llamaopenefficientfoundation} have exhibited impressive ability to understand and generate human language, making them highly applicable to AC \cite{Zhang2024AffectiveCI, Zhang2023SentimentAI, sun2023sentimentanalysisllmnegotiations, taffco/AminMCS24, zhao-etal-2024-esc}. LLMs operate on the principle of in-context learning (ICL) \cite{Brown2020LanguageMA, nips/AnML0LC24, Kojima_Shixiang_Gu_Reid_Matsuo_Iwasawa, Wei2021FinetunedLM, Zhang_Zhang_Li_Smola_2022}, allowing them to generate coherent and contextually appropriate text based on prompts or preceding content. When scaled appropriately, LLMs exhibit emergent reasoning abilities, particularly in solving complex tasks through chain-of-thought (CoT) prompting \cite{Wei2022ChainOT, Kojima_Shixiang_Gu_Reid_Matsuo_Iwasawa}. This capability has critical implications for AC tasks, as it enables models to dynamically adapt their responses to the emotional context of interactions. For instance, LLMs can leverage CoT and ICL to infer user emotions and generate responses that are contextually sensitive and emotionally relevant \cite{Zhang2023SentimentAI, xu2024ds2absadualstream, cikm/LiuZLZC24}.  Most existing LLMs are based on the Transformer \cite{Vaswani2017AttentionIA} architecture and can be categorized into three types: Encoder-only (e.g., BERT \cite{Devlin_Chang_Lee_Toutanova_2019}, Decoder-only (e.g., CHATGPT \cite{CHATGPT} and Encoder-Decoder (e.g, Flan-T5 \cite{flant5}) models. Research has explored collaboration among these different architectures to enhance the capabilities of LLMs on downstream tasks through fine-tuning \cite{HoSY23, HsiehLYNFRKLP23} or multi-task learning \cite{rvisa, Explanations}. Given the computational cost associated with full parameter training, parameter-efficient fine-tuning methods, such as LoRA \cite{Hu_Shen_Wallis_Allen-Zhu_Li_Wang_Chen_2021}, or speculative decoding \cite{icml/LeviathanKM23, acl/XiaYDW00L0S24} methods, have been proposed to enable efficient training and inference.

Despite these advancements, LLMs still struggle with cognitive challenges in affective reasoning, such as hallucinations \cite{Zhang2023SirensSI, Li2024ASO}, misinterpretation of cultural nuances, and failure to recognize subtleties in emotional contexts \cite{emnlp/AdilazuardaMLSA24, nips/LiC0S024}. These issues are particularly pertinent in diverse settings where emotional expressions can vary significantly across cultures and contexts. Recent research indicates a growing need for LLM-based collaborative systems \cite{zhang2024collaborativeaisentimentanalysis, HeCollaborative, Schleiger31122024} to address these challenges. Such systems can integrate specialized models focused on emotional intelligence and contextual understanding, combining their strengths to emulate human-like affective reasoning. By fusing the capabilities of LLMs with those of specialized models and fostering collaboration among LLMs themselves, we can develop systems that exhibit greater robustness and adaptability in addressing complex emotional interactions, paving the way for more natural and effective human-computer collaboration.

%% file: src/collaboration_framework.tex
\section{LLM-based Collaboration Systems}
\label{framework}
\subsection{Collaboration Strategies}
\begin{figure*}[h!]
\centering
\includegraphics[width=0.8\textwidth]{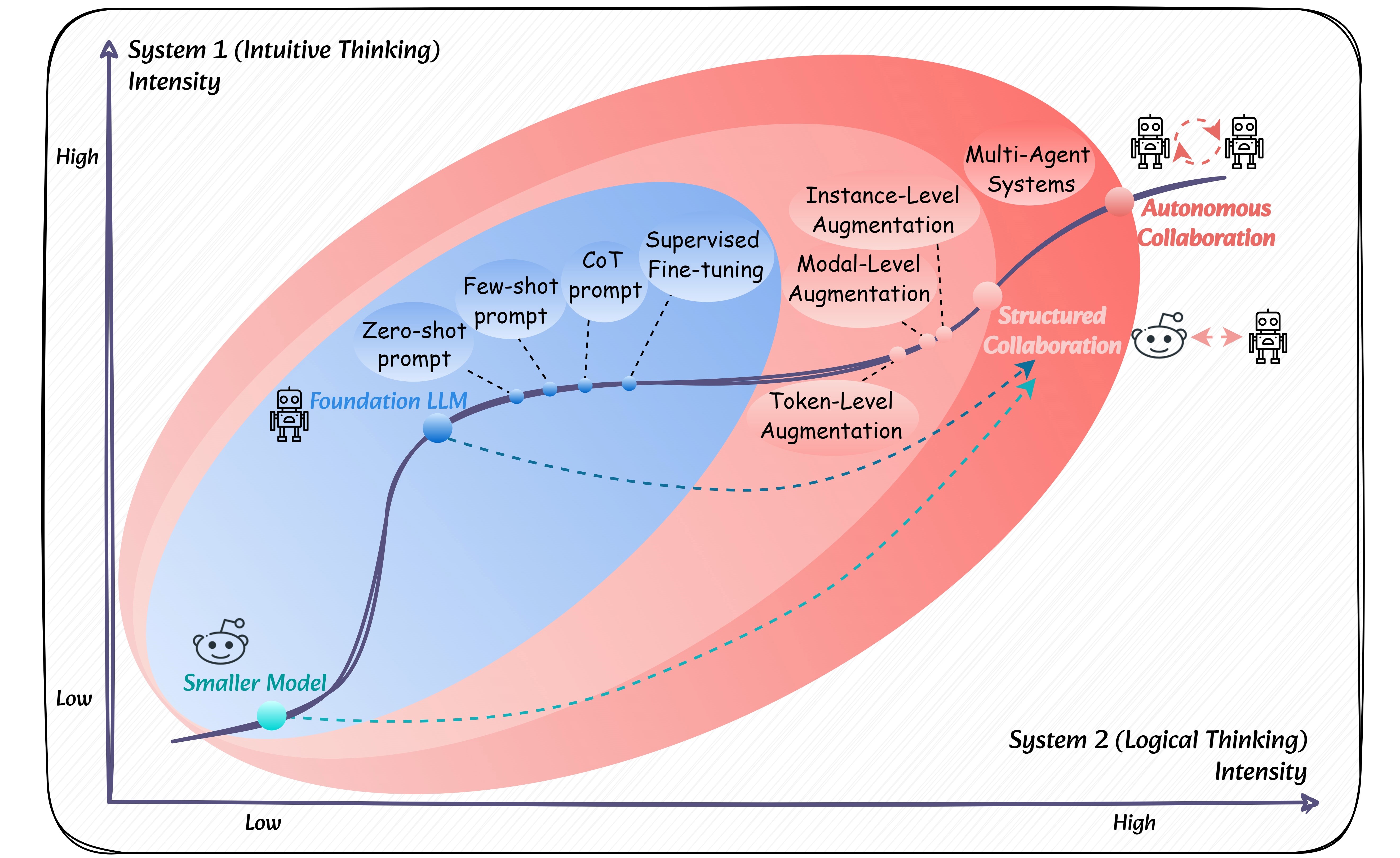}
\caption{Development of LLM-based collaboration, regarding the intensity of system 1 (intuitive thinking) and system 2 (logical thinking).}
\label{fig:comparison}
\end{figure*}

LLM-based collaboration strategies can be categorized into structured and autonomous collaboration according to model interaction patterns, as illustrated in Figure \ref{fig:framework}. These strategies are aligned with cognitive processes supported by \emph{Dual Process Theory}. As we progress from smaller models to foundational LLMs, there is a significant enhancement in fast, intuitive thinking across various downstream tasks. Techniques such as few-shot prompting, Chain of Thought (CoT), and supervised fine-tuning enable foundational LLMs to engage in slower, more logical thinking for complex reasoning tasks \cite{Brown2020LanguageMA, HoSY23, Wei2022ChainOT, Kojima_Shixiang_Gu_Reid_Matsuo_Iwasawa}. Structured collaboration coordinated the strengths of specialized models and LLMs to enhance both system 1 (intuitive) and system 2 (logical) thinking concurrently. With the development of more advanced LLMs, multi-agent systems can facilitate automatic and iterative interactions. Figure \ref{fig:comparison} demonstrates the trend towards increasingly dynamic and socially aware LLM-based collaboration systems, driven by advancements in synchronizing cognitive processes. In the following sections, we will provide an overview of the differences between structured and autonomous collaboration, followed by an in-depth exploration of the strategies within these two categories.

% Table \ref{collaboration trend} contrasts these paradigms in terms of model roles and data flow, highlighting a trend toward increasingly dynamic and socially aware systems, driven by advancements in data evolution and cognitive process synchronization.

\textbf{Structured Collaboration} employs a hierarchical division of labor, where different models assume distinct roles to achieve a common goal. Drawing an analogy to Kahneman's \emph{System 1}, models such as emotion recognizers or intent classifiers process inputs in real-time using task-specific knowledge. They are designed to quickly and intuitively handle specific aspects of the task, providing immediate outputs that are then utilized by backbone models, which are akin to \emph{System 2}. The backbone models engage in slower, more deliberate reasoning processes, such as generating a response conditioned on the detected emotion. This one-way data flow ensures controllability on model composition and task precision, as each model operates within its predefined scope, contributing to the overall task in a coordinated manner.

\textbf{Autonomous Collaboration}, in contrast to structured collaboration, empowers LLMs to function as proactive agents with either shared or competing objectives. This paradigm leverages the flexibility and adaptability of LLMs to operate in dynamic environments, where agents engage in perception-feedback loops, such as iterative negotiation protocols. These iterative loops enable agents to continuously adjust their roles and strategies, balancing slow and fast thinking within each agent. For example, an agent might first assess the perceived affective tone (acting as a discriminator with fast thinking ), then receive interaction information and generate self-reflective analysis (acting as a generator with slow thinking). This autonomy and flexibility allow agents to respond to complex, open-ended scenarios with a high degree of adaptability. However, the inherent complexity of autonomous multi-agent interaction necessitates careful design and management to prevent chaotic behavior. Reward shaping becomes a critical component in guiding agent behavior, ensuring that agents align with desired outcomes and collaborate effectively. 

\input{src/taxonomy}

% \begin{table*}[h!t]
%   \caption{Comparison of LLM-based collaboration strategies, emphasizing the transition from static to dynamic and socially aware systems.}
%   \label{collaboration trend}
%   \centering
%   \begin{tabular}{p{3.5cm}p{6.5cm}p{6.5cm}}
%     \toprule
%      & Structured Collaboration & Autonomous Collaboration\\
%     \midrule
%     \multirow{2}{*}{\hfil Model Roles} & Static division & Dynamic adjustment \\
%    & Distinct roles for System 1 (fast) and System 2 (slow) &  Agents capable of flexible System 1 and 2 thinking \\
%    \midrule
%     % Environment Perception  & Context window & Real-time feedback loops \\
%     Data Flow & Unidirectional & Bidirectional / Iterative \\
%     % Knowledge Update & Frozen post-training & Online adaptation \\
%     \bottomrule
%   \end{tabular}
% \end{table*}

Given an affective dataset $D = \{x, y\}^N$ containing inputs $x_i = x_1...x_n, i \in [1,N]$, both affective understanding and generation tasks can be regarded as language generation tasks, with outputs $y_i = y_1...y_n, i \in [1, N]$, irrespective of whether they involve label prediction or sequence generation. We classify the following collaboration strategies focusing on mechanisms, key functions, and applications.
\subsection{Structured Collaboration}
\label{Sturctured}
\subsubsection{Token-level Augmentation}
\label{TOKEN}
\begin{figure*}[h!]
\centering
\includegraphics[width=\textwidth]{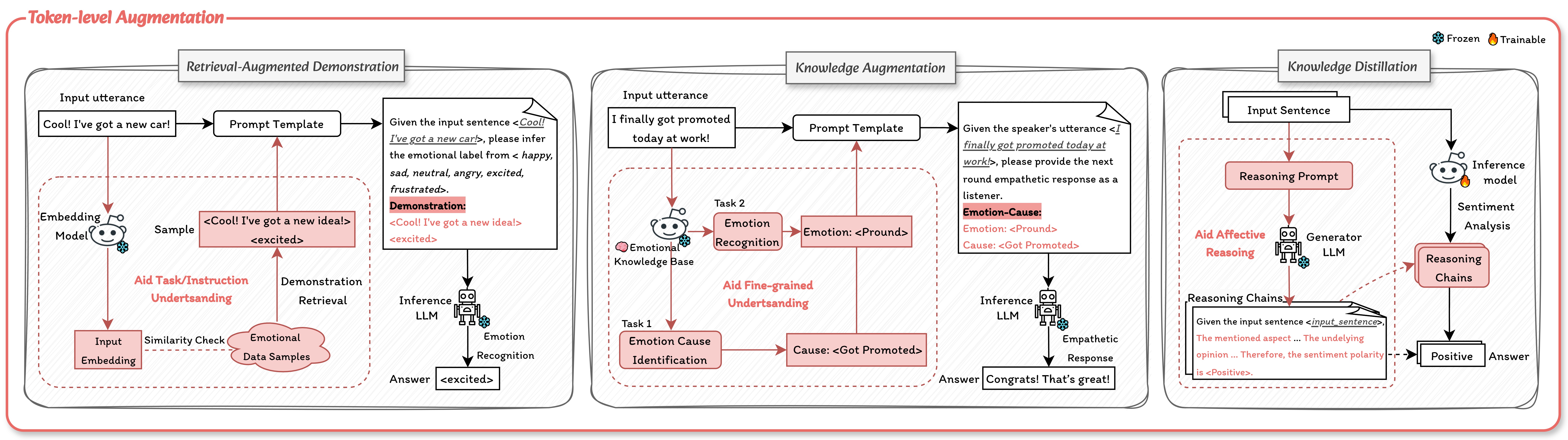}
\caption{Token-level augmentation strategies in structured collaboration. Three strategies are illustrated with examples, including Retrieval-Augmented Demonstration (\textsection~\ref{RAD}), Knowledge Augmentation (\textsection~\ref{KA}), and Knowledge Distillation (\textsection~\ref{KD}).}
\label{fig:token-level}
\end{figure*}
Token-level augmentation enhances input or output sequences by injecting task-specific signals (e.g., demonstrations, knowledge snippets) into the token stream. Figure \ref{fig:token-level} illustrates three dominant methods: Retrieval-Augmented Demonstration, Knowledge Augmentation, and Knowledge Distillation.
\paragraph{Retrieval-Augmented Demonstration}
\label{RAD}
\begin{itemize}
    \item \textbf{Mechanism:} Using embedding models for sentence similarity checks to retrieve relevant context or demonstrations $\mathcal{D}(x)$ from data source $\mathcal{S}$, then injecting them into the input context: 
    \begin{equation}
    \begin{aligned}
        x'= x \oplus \mathcal{D}(x), \qquad \qquad \qquad\\ 
        \mathcal{D}(x) = \text{Top-k}\,\text{argmax}_{d \in \mathcal{S}} \text{Sim}(E(x), E(d)) 
    \end{aligned}
    \end{equation}
    where $E(\cdot)$ denotes embedding vectors and $\text{Sim}(\cdot)$ denotes similarity calculation.
    \item \textbf{Key Function:} This approach enhances task or instruction understanding and improves context-aware affective reasoning by grounding generation in contextually relevant scenarios.
    \item \textbf{Applications:} In affective understanding, Sun et al. \cite{sun2023sentimentanalysisllmnegotiations} applied RoBERTa-Large \cite{Liu2019RoBERTaAR} to retrieve the k most similar neighbors as demonstrations in prompts, enabling zero-shot sentiment classification via in-context learning. Similarly, InstructERC \cite{lei2024instructerc} designed a retrieval module using Sentence-BERT \cite{sentence-bert} to extract the most similar examples from training data and enrich the prompt context along with instruction, historical content, and label statement for the ERC task. This strategy relies solely on input text $x$, with retrieved demonstrations closely aligned with target sentences, thereby improving task inference on analogous cases by capitalizing on the generalization capabilities of LLMs.  In affective generation, Cue-CoT \cite{wang-etal-2023-cue} developed Bert \cite{Devlin_Chang_Lee_Toutanova_2019} as the embedding model to find nearest examples along with Chain-of-Thought prompting, eliciting empathetic reasoning in LLMs for empathic response generation. Another research study \cite{Liu2024SpeakFromHeart} adopted contrastive learning to develop an initial embedding model that captures emotion-enhanced representations for input utterances by considering emotion labels. The trained model is then applied in subsequent emotional demonstration retrieval. They not only considered demonstration support and historical context but also incorporated multimodal information, such as video clip captions, to enrich the emotional context. 
    Across both affective understanding and generation tasks, these applications share a common paradigm of demonstration selection facilitated by embedding models, with LLMs serving as the backbone for primary reasoning. Both tasks are treated as generative, with no significant differences in method design. However, the success of these applications largely depends on the embedding model's ability to capture nuanced emotional semantics, which is crucial for accurately interpreting and generating affective content. \\
\end{itemize}

\paragraph{Knowledge Augmentation}
\label{KA}
\begin{itemize}
    \item \textbf{Mechanism:} Using a specialized model $\mathcal{M}(\cdot)$ as the emotional knowledge base or tool to provide commonsense or domain-specific knowledge by curating knowledge data $\mathcal{K}(x)$, which is concatenated to the input:
    \begin{equation}
        x' = x \oplus \mathcal{K}(x), \quad \mathcal{K}(x) = \mathcal{M}(x)
    \end{equation}
    \item \textbf{Key Function:} This approach focuses on external knowledge injection by injecting emotional knowledge (e.g., rules, facts, heuristics) to compensate for limited affective grounding and enhance fine-grained understanding.
    \item \textbf{Applications:} In contrast to demonstration augmentation, knowledge-guided augmentation enriches emotional understanding by integrating knowledge from diverse sources. Models trained on domain-specific datasets can offer valuable insights pertinent to that domain, while LLMs, renowned for their ability to handle multiple tasks and trained on extensive corpora, are adept at providing commonsense knowledge. 
    Specifically, \cite{cikm/LiuZLZC24} utilized heterogeneous LLMs to simultaneously provide linguistic, background, and commonsense knowledge, which were then fused to create a zero-shot reasoning context for an off-the-shelf LLM in aspect-based sentiment analysis tasks. To guide affective generation with emotional nuances, HEF \cite{yang2024enhancingerg} trained a small-scale empathetic model to provide emotion categories and emotional cause words, which were then used to guide LLM instructions for empathetic response generation. In contrast, a recent work \cite{cao-etal-2025-tool} leveraged domain-expert models as emotional knowledge tools. They recorded tool call behaviors to develop a tool-learning paradigm for empathetic response generation, enhancing the backbone model ability to request additional tools and incorporate appropriate knowledge effectively. Another research study \cite{bhaskar-etal-2023-prompted} explored topic clustering and aspect extraction through a supervised learning model, integrating the obtained knowledge into LLM instructions to focus on salient content summarization. Sibyl \cite{wang-etal-2025-sibyl} investigated visionary commonsense knowledge by training an LLaMA \cite{llama2023} model to deduce four categories of commonsense knowledge. This knowledge was then used to guide task-solver fine-tuning, improving the ability to apply visionary commonsense reasoning in various contexts. 
    Overall, these applications enhance the model capacity to understand and generate emotionally aware content by leveraging external knowledge sources. However, it is crucial to carefully verify the quality and accuracy of both the knowledge provider and the knowledge itself.
\end{itemize}

\paragraph{Knowledge Distillation}
\label{KD}
\begin{itemize}
    \item \textbf{Mechanism:} Using a teacher model $\text{Teacher}(\cdot)$ to generate reasoning chains or internal knowledge $\mathcal{R}(x,y)$ conducive to task inference, then guiding the student model $Student(\cdot)$ to mimic the inference process.
    \begin{equation}
        y' = y \oplus \mathcal{R}(x,y),\quad \mathcal{R}(x, y) = \text{Teacher}(x, y)
    \end{equation}
    \item \textbf{Key Function:} This approach aims to guide student models to emulate the problem-solving process of the teacher model, balancing effectiveness and efficiency while preserving interpretability.
    \item \textbf{Applications:} Compared to the other two methods of token-level augmentation, knowledge distillation involves mimicking the inference process of more powerful LLMs, which serve as teacher models. This approach enhances the grounding information in the output response by transferring knowledge from the teacher to a smaller student model. For example, SCRAP \cite{kim-etal-2024-self-consistent} distilled the T5 model \cite{Raffel2019ExploringTL} for aspect sentiment quad prediction by prompting GPT-3.5-turbo to generate reasoning chains as groundings along with ground truth labels. Similarly, another research study \cite{zhang2024distillingfinegrained} designed analysis and rewriting prompts to expand the output response of two additional datasets from the teacher model, then distilling fine-grained sentiment understanding ability by focusing on analysis and rewriting perspectives. In contrast, Sibyl \cite{wang-etal-2025-sibyl} generated grounding visionary commonsense knowledge from the teacher model and guided the student model in developing the ability to generate grounding knowledge through fine-tuning. Additionally, Cao et al. \cite{CaoImprovingESC} employed powerful LLMs like GPT-3.5 to provide more practical strategy-intent knowledge for emotional support conversations. This knowledge assisted in fine-tuning the model to select appropriate strategies and clarify the intent behind them during output generation.
    In general, an LLM usually serves as the teacher model to generate step-by-step rationales or common knowledge. The smaller student model is trained on both the generated groundings and final answers, enabling it to internalize the reasoning logic or knowledge effectively. This process not only improves the student model performance but also reduces its dependency on large-scale computational resources, making it more efficient for practical applications. However, careful curation of teacher outputs is required to avoid propagating biases.
\end{itemize}

\subsubsection{Instance-level Augmentation}
\label{INSTANCE}
\begin{figure*}[h!]
\centering
\includegraphics[width=\textwidth]{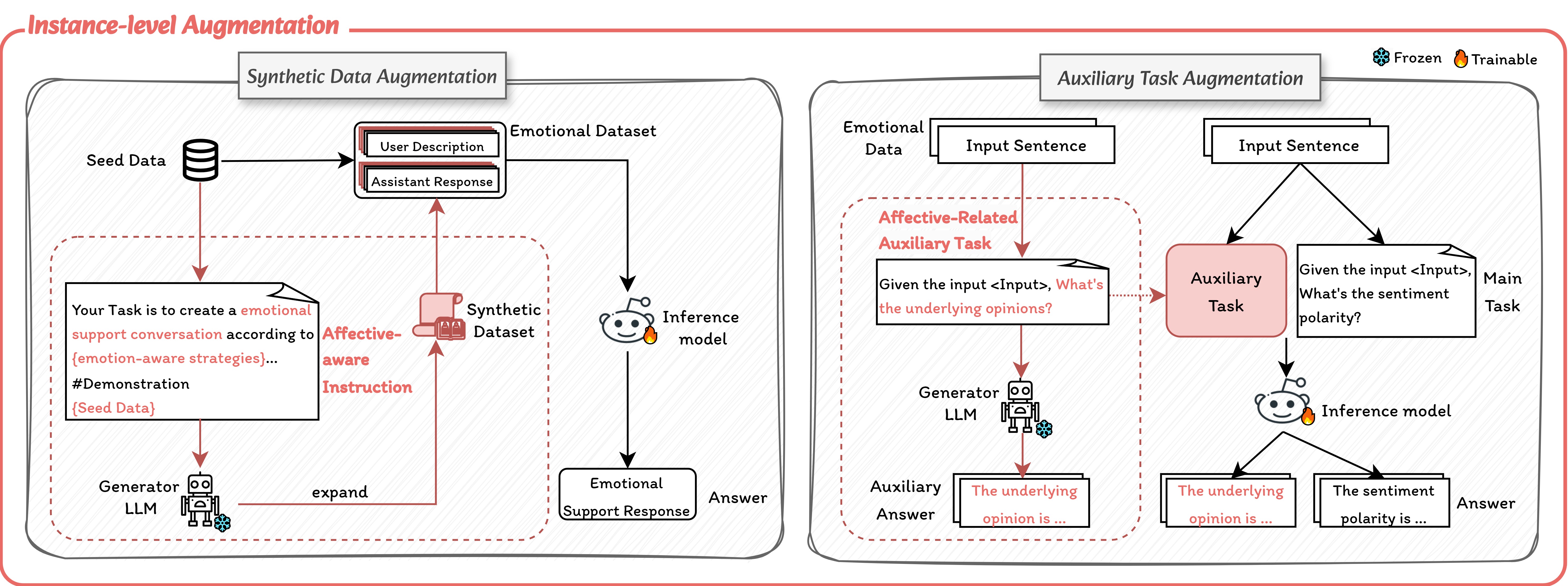}
\caption{Instance-level augmentation strategies in structured collaboration. Two strategies are illustrated with examples, including Synthetic Data Augmentation (\textsection~\ref{SDA}) and Auxiliary Task Augmentation (\textsection~\ref{ATA}).}
\label{fig:instance-level}
\end{figure*}
Instance-level augmentation operates on entire data samples rather than individual tokens. It is pivotal in addressing data scarcity and improving model performance in affective computing tasks. Key strategies include Synthetic Data Augmentation and Auxiliary Task Augmentation, as illustrated in Figure \ref{fig:instance-level}.

\paragraph{Synthetic Data Augmentation}
\label{SDA}
\begin{itemize}
    \item \textbf{Mechanism:} Using LLMs to generate synthetic data samples $\mathcal{S}$ based on the existing dataset $\mathcal{D}$, the augmented dataset $\mathcal{D}'$ is then employed to train the backbone model:
    \begin{equation}
        \mathcal{D}' = \mathcal{D} \cup \mathcal{S} = {(x_i, y_i)}_{i=1}^{N+k}, \quad \mathcal{S} = \text{LLMs}(\mathcal{D}) 
        % = {(x_j, y_j)}_{j=1}^{k} 
    \end{equation}
    
    \item \textbf{Key Function:} This approach enhances data quantity and diversity, effectively addressing the challenge of data scarcity in affective computing.
    \item \textbf{Applications:} In affective understanding, various studies have explored synthetic data augmentation. For instance, this study \cite{xu2024ds2absadualstream} employed dual-stream data synthesis for ABSA tasks. The process involves prompting LLMs to perform conditional generation by brainstorming relevant attributes and applying multi-level transformations to seed data, thereby increasing data diversity and improving model training efficacy. Another study \cite{XuXQWT25} focused on altering sentiment polarity using detailed LLM instructions, which aids in capturing sentiment nuances. In contrast, this method \cite{zhong2024iterative} proposed a three-stage synthetic data generation framework to extract and extend aspects, focusing on generating data from diverse aspects and further enhancing the performance of existing training methods.
    While in affective generation tasks, SoulChat \cite{chen-etal-2023-soulchat} developed a self-curated dataset by converting single-turn long-text psychological counseling conversations into multi-turn empathy conversations, with manual proofreading and data cleansing to strengthen empathy expression. Zheng et al. \cite{zheng-etal-2024-self} generated an augmented dataset by iteratively prompting LLMs to expand conversation samples based on comprehensive scenarios and strategies. Recent advancements include collaborative agent interactions to simulate human behavior with sticker usage, creating a new benchmark dataset for empathetic responses accompanied by sticker representation \cite{zhang-etal-2024-stickerconv}. 
    These diverse methods, ranging from manual design to autonomous LLM interactions, enhance instance diversity and quantity, facilitating more effective model training for challenging tasks like ABSA and empathetic response generation. 
\end{itemize}

\paragraph{Auxiliary Task Augmentation}
\label{ATA}
\begin{itemize}
    \item \textbf{Mechanism:} Using LLM to augment auxiliary task data $\mathcal{D}_{aux}$ given primary task $\mathcal{D}_{\text{pri}}$, it usually minimizes a combined loss function to train the backbone model jointly:
    \begin{equation}
    \mathcal{D}' = \mathcal{D}_{\text{pri}} \oplus  \sum_{i=1}^{m} \mathcal{D}_{\text{aux}}^i, \quad \mathcal{D}_{\text{aux}}^i = \text{LLMs}(\mathcal{D}_{\text{pri}})
    \end{equation}
    where $m$ denotes the number of auxiliary tasks.
    \item \textbf{Key Function:} This approach aims to augment the primary task with auxiliary tasks that are affective-related or task-relevant, thereby reinforcing primary task learning.
    \item \textbf{Applications:} Auxiliary task augmentation is prevalent in affective understanding tasks, where it enhances the robustness of nuanced emotion perception. For example, RVISA \cite{rvisa} designed reasoning and verification tasks for implicit sentiment analysis, mimicking human reasoning processes with tasks reinforcing each other. MT-ISA \cite{lai2024mtisa} proposed supplementing sentiment elements as auxiliary tasks with multi-task learning, where sentiment elements enrich each other for sentiment polarity reasoning. In the emotional recognition task, InstructERC \cite{lei2024instructerc} developed a speaker identification task to focus on targeted emotion impact prediction, training incrementally from sub-task to main task. ECR-Chain \cite{ijcai2024ECRChain} developed a reasoning task from LLMs and trained jointly with a primary emotion prediction task to enhance reasoning learning. CKERC \cite{fu2024ckercjointlarge} instead developed a commonsense identification task to aid emotion prediction, trained jointly in a pipeline. Despite these, auxiliary tasks can enhance multimodal representation learning in emotional recognition by pre-training on auxiliary data from additional modalities \cite{dutta2025llmsupervisedpretrainingmultimodal}. 
    Overall, auxiliary tasks serve as additional signals that can constrain and enhance the primary task through joint learning, playing a crucial role in mutual reinforcement within the loss function. However, they often emphasize localized, classification-driven objectives, which can introduce conflicting optimization signals that may suppress the creative or nuanced expression required in affective generation tasks. These tasks prioritize open-ended, contextually coherent outputs that rely on flexible, holistic reasoning rather than rigid, structured predictions. Consequently, the utility of auxiliary tasks in affective generation is limited due to these divergent task objectives.
\end{itemize}

\subsubsection{Modal-level Augmentation}
\label{MODAL}
\begin{figure*}[h!t]
\centering
\includegraphics[width=\textwidth]{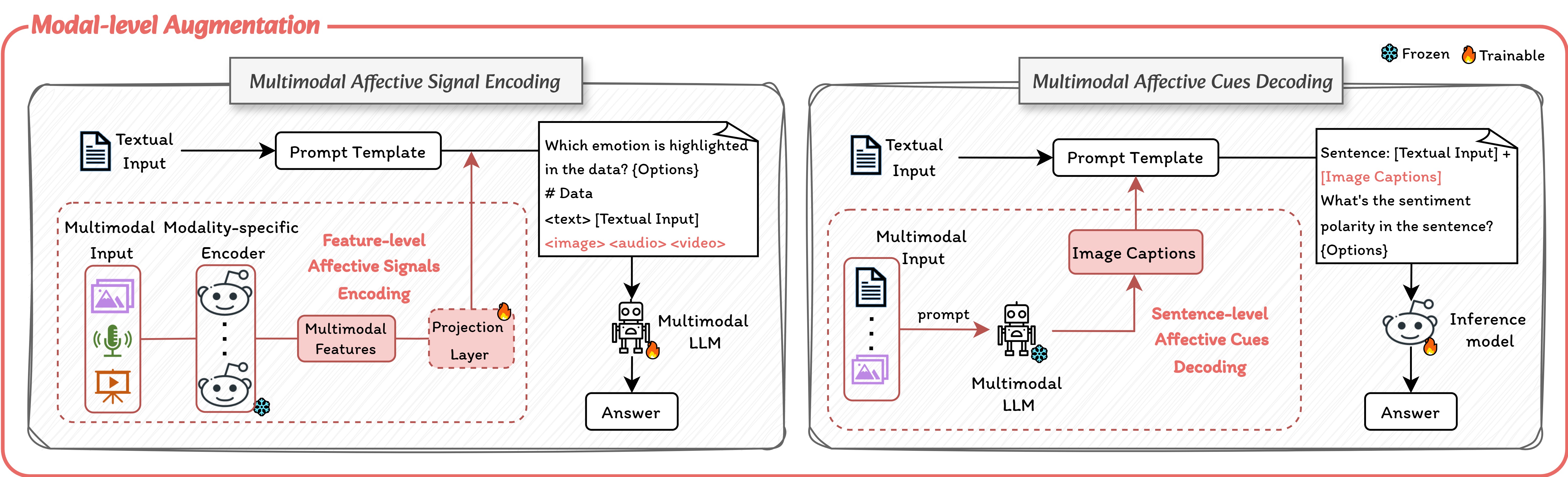}
\caption{Modal-level augmentation strategies in structured collaboration. Two strategies are illustrated with examples, including Multimodal Affective Signal Encoding (\textsection~\ref{M-encoding}) and Multimodal Affective Cues Decoding (\textsection~\ref{M-decoding}).}
\label{fig:modal-level}
\end{figure*}
Modal-level augmentation leverages multimodal capabilities in advanced LLMs to unify heterogeneous affective signals (text, audio, visual) under a cohesive framework. Key strategies include Multimodal Affective Signal Encoding and Multimodal Affective Cues Decoding, as illustrated in Figure \ref{fig:modal-level}, addressing the dual challenges of cross-modal alignment and affective context preservation.
\paragraph{Multimodal Affective Signal Encoding}
\label{M-encoding}
\begin{itemize}
    \item \textbf{Mechanism:} Extracting multimodal features $h(\cdot)$ through modality-specific encoders $Encode(\cdot)$ and fusing these features through LLMs to achieve holistic emotion perception.:
    \begin{equation}
        \mathcal{D}' = \mathcal{D} \oplus h(\mathcal{D}_m), \quad h(\mathcal{D}_m) = Encode(\mathcal{D}_m)
    \end{equation}
    where $m$ denotes modalities including audio, image, video, etc.
    \item \textbf{Key Function:} This approach focuses on bridging different modalities by incorporating lightweight preprocessing tools, enabling a comprehensive understanding of emotional signals across various forms of data.
    \item \textbf{Applications:} In affective understanding, EmoLLM \cite{yang2024emollm} transformed multimodal inputs into feature vectors using a projection layer, allowing LLMs to process them effectively. This was followed by LoRA fine-tuning \cite{Hu_Shen_Wallis_Allen-Zhu_Li_Wang_Chen_2021} on a self-curated emotional understanding dataset, along with a generated reasoning chain, to enhance affective understanding. In affective generation, STICKERCONV \cite{zhang-etal-2024-stickerconv} explored a similar design by utilizing a frozen modality-specific encoder alongside a trainable linear layer to encode images into aligned feature vectors. These vectors were then fed into LLMs with LoRA fine-tuning, facilitating empathetic response and sticker generation. ELMD \cite{Liu2024SpeakFromHeart} focused on video data, applying temporal and spatial feature-level encoding with a trainable linear layer. This approach allowed for effective fine-tuning by LLMs, enabling them to process complex video features for improved emotion recognition. These methods collectively highlight the power of integrating multimodal data to enhance affective computing tasks. 
    By transforming diverse inputs into a learnable feature space adaptive to backbone models, these approaches enable LLMs to better interpret and respond to multimodal emotional signals. The use of LoRA fine-tuning across different modalities demonstrates its effectiveness in improving model performance, allowing for more nuanced and contextually aware interactions. However, it is still challenging to generalize across backbone models and ensure different modalities are properly aligned for interpretation.
\end{itemize}

\paragraph{Multimodal Affective Cues Decoding}
\label{M-decoding}
\begin{itemize}
    \item \textbf{Mechanism:} Decoding multimodal input into affective cues in text format $\mathcal{T}(\cdot)$, which are then incorporated into context information for subsequent training or inference. 
    \begin{equation}
         \mathcal{D}' = \mathcal{D} \oplus \mathcal{T}(\mathcal{D}_m), \quad \mathcal{T}(\mathcal{D}_m) = Decode(\mathcal{D}_m)
    \end{equation}
    \item \textbf{Key Function:} This approach focuses on bridging modalities by converting multimodal data into textual affective cues, facilitating the integration of diverse emotional signals into a coherent context for LLM processing.
    \item \textbf{Applications:} In affective understanding, DialogueLLM \cite{zhang2024dialoguellm} utilized a multimodal LLM to convert video data into captions, which were incorporated into instructions for LLaMA fine-tuning. This augmented video context, enriched with affective cues, enhances emotion recognition in conversations. Similarly, Feng et al. \cite{ feng-etal-2024-affect} employed a lightweight audio speech recognition model to transform speech into transcripts, integrating them into prompts alongside text samples. They conducted LoRA fine-tuning on LLaMA to evaluate the ability of LLM in handling potential audio recognition errors. Wisdom \cite{mm/Wang0S00T24} a vision-language model to generate contextual information from images and sentences, using a rule-based fusion mechanism to adapt this knowledge for multimodal sentiment analysis. In affective generation,  SMES \cite{chu2024multimodalemo} used Video-Llama to extract emotional cues from video and audio, processing them through fine-tuning on an LLM-based reasoning model with sequential training from subtasks to emotional support response generation. These methods collectively demonstrate the potential of multimodal affective cues decoding approaches in advancing affective computing tasks. 
    By leveraging cues from multiple modalities, these systems can generate responses that are both contextually aware and emotionally resonant. This comprehensive integration enables a deeper understanding of emotional contexts, facilitating more appropriate responses to multimodal inputs. Although some noise may be introduced during the transformation into text, exploring more efficient methods for real-time systems remains an area for further research and development.
\end{itemize}

\subsection{Autonomous Collaboration}
\label{AUTO}

\subsubsection{Multi-agent Systems}
\label{MAS}
\begin{figure*}[h!t]
\centering
\includegraphics[width=\textwidth]{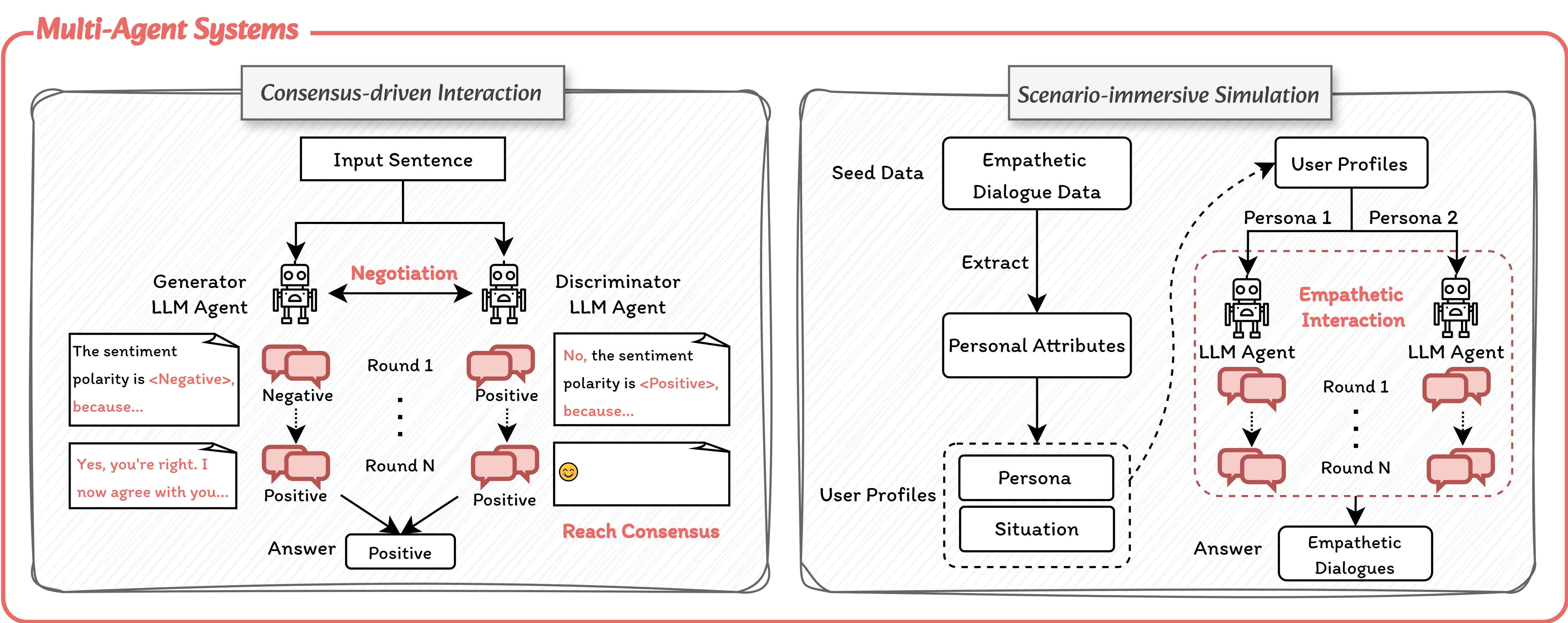}
\caption{Multi-agent systems for autonomous collaboration. Two strategies are illustrated with examples, including Consensus-driven Interaction (\textsection~\ref{CDI}) and Scenario-immersive Simulation (\textsection~\ref{SIS}).}
\label{fig:MAS}
\end{figure*}
Multi-agent systems leverage the collective intelligence of LLM-based agents to tackle complex affective tasks through emergent social behaviors. Key strategies include Consensus-driven Interaction for structured decision-making and Scenario-immersive Simulation for open-ended emotional dynamics, as illustrated in Figure \ref{fig:MAS}.

\paragraph{Consensus-driven Interaction}
\label{CDI}
\begin{itemize}
    \item \textbf{Mechanism:} Multiple agents $\mathcal{A}_i$ engage in debate, negotiation, or voting to exchange ideas and eventually reach a consensus or agreement on affective decisions:
    \begin{equation}
    \begin{split}
        y_{\text{predict}} = \sum_{i=1}^n w_i \cdot \mathcal{A}_i(x; h_{t-1}), \\ \quad h_{t-1} \sim  \mathcal{H}(s_{t-1}, a_{t-1}, \ldots)
    \end{split}
    \end{equation}
    where $w_i$ denotes the voting weight and $h_{t-1}$ is modeled as a function $\mathcal{H}$ that captures relevant historical data, such as the previous state $s_{t-1}$, actions $a_{t-1}$, and potentially other past interactions.
    \item \textbf{Key Function:} This approach aggregates diverse perspectives to reduce individual biases, enhancing robustness in affective decision-making.
    \item \textbf{Applications:}  The consensus-driven approach tends to be more practical for tasks that require precise answers. For instance, \cite{sun2023sentimentanalysisllmnegotiations} evaluated multi-agent negotiations on sentiment analysis, finding that role-flipping multi-LLM negotiation methods surpass traditional in-context learning methods. This highlights the potential of leveraging diverse perspectives from multiple agents to enhance decision-making accuracy. However, gaps remain compared to some supervised learning methods. Similarly, HAD \cite{FrankAgentforSA} explored the use of heterogeneous LLMs for negotiations in financial sentiment analysis, rooted in Minsky’s theory of mind and emotions. Their approach achieved better accuracies compared to alternative multi-LLM settings, particularly when the discussion content was substantial.  
    In general, consensus-driven interaction leverages the diverse viewpoints of multiple agents to converge on optimal solutions. This is beneficial in domains or tasks where nuanced understanding and agreement are essential. Considering the effect associated with introducing diverse LLM-based agents, there is a trade-off between diversity and cost efficiency.
\end{itemize}

\paragraph{Scenario-immersive Simulation}
\label{SIS}
\begin{itemize}
    \item \textbf{Mechanism:} Multiple agents $\mathcal{A}_i$ with predefined role attributes $r_i$ dynamically interact in state-evolving environments:
    \begin{equation}
    \begin{split}
        y_{\text{predict}}  = s_{t+1} = f(s_t, {a_i}_{i=1}^n), \quad a_i \sim \mathcal{A}_i(x, s_t; r_i)
    \end{split}
    \end{equation}
    where $s_t$ denotes the dialogue state at turn $t$, $n$ is the number of agents, and $f(\cdot)$ models social dynamics of agents.
    \item \textbf{Key Function:} This approach adaptively generates emotionally engaging and contextually coherent interactions by embedding role-specific attributes. This mimics social dynamics while enhancing creativity and diversity.
    \item \textbf{Applications:} For affective generation tasks requiring creative and emotionally-aware outputs, scenario-immersive simulation through multi-agent interaction is effective in dynamically expanding topics and evolving diverse, contextually-aware behaviors. STICKERCONV \cite{zhang-etal-2024-stickerconv} applied agent frameworks for empathetic dialogue generation with stickers, mimicking human conversational patterns and ensuring contextual consistency and variety. Cooper \cite{aaai/ChengLWLOLW024} explored dividing emotional support conversation goals into multiple aspects for specialized agent modules, where each module dynamically adjusts its progress state and facilitates the next round of topic selection. VLESA \cite{li-etal-2024-helpful} proposed optimal relevance learning inspired by the cognitive relevance principle, simulating user and assistant interactions with a helpfulness scorer to provide feedback and facilitate aligned behaviors. 
    These applications underscore that scenario-immersive simulation excels in generating creative and emotionally-rich interactions, allowing agents to adapt and respond to dynamic contexts with empathy and relevance. However, the detailed division of tasks may increase computational demands due to iterative interactions and updates.
\end{itemize}

%% file: src/taxonomy.tex
\begin{figure*}[h!t]
\centering
\resizebox{\linewidth}{!}{
\tikzset{
        my node/.style={
            draw,
            align=center,
            thin,
            text width=1.2cm, 
            rounded corners=2,
        },
        my leaf/.style={
            draw,
            align=left,
            thin,
            text width=8.5cm, 
            % text height=1cm, 
            % minimum height=0.5cm,
            rounded corners=2,
        }
}
\forestset{
  every leaf node/.style={
    if n children=0{#1}{}
  },
  every tree node/.style={
    if n children=0{minimum width=1em}{#1}
  },
}
\begin{forest}
    nonleaf/.style={font=\scriptsize},
     for tree={%
        % my node,
        every leaf node={my leaf, font=\tiny},
        every tree node={my node, font=\tiny, l sep-=4.5pt, l-=1.pt},
        anchor=west,
        inner sep=2pt,
        % l = 10pt,
        l sep=10pt, % control leaf to parent nodes gaps (horizontal)
        s sep=4mm, % control node gaps (vertical)
        fit=tight,
        grow'=east,
        edge={ultra thin},
        parent anchor=east,
        child anchor=west,
        if n children=0{}{nonleaf}, 
        edge path={
            \noexpand\path [draw, \forestoption{edge}] (!u.parent anchor) -- +(5pt,0) |- (.child anchor)\forestoption{edge label};
        },
        if={isodd(n_children())}{
            for children={
                if={equal(n,(n_children("!u")+1)/2)}{calign with current}{}
            }
        }{}
    }
    [\rotatebox{90}{\textbf{LLM-based Collaboration Strategies}}, draw=gray, fill=gray!10, text width=1.2em, text=black,
        [\textbf{Structured Collaboration}, color=pink, fill=pink!60, text width=6em, text=black
            [\textbf{Token-level Augmentation} (\textsection~\ref{TOKEN}), color=pink, fill=pink!60, text width=6em, text=black
                [Retrieval-Augmented Demonstration, color=pink, fill=pink!60, text width=6em, text=black
                    [AU, color=blue!40, fill=blue!20, text width=1.2em, text=black [ELMD \cite{Liu2024SpeakFromHeart}; Cue-CoT \cite{wang-etal-2023-cue}, color=pink!40!blue!80!red!40, fill=pink!10!blue!80!red!10, text width=3cm, text=black]]
                    [AG, color=blue!40, fill=blue!20, text width=1.2em, text=black [Negotiation \cite{sun2023sentimentanalysisllmnegotiations}; InstructERC \cite{lei2024instructerc}, color=pink!40!blue!80!red!40, fill=pink!10!blue!80!red!10, text width=3cm, text=black]]
                ]
                [Knowledge Augmentation, color=pink, fill=pink!60, text width=6em, text=black
                    [AU, color=blue!40, fill=blue!20, text width=1.2em, text=black [MLKF \cite{cikm/LiuZLZC24}, color=pink!40!blue!80!red!40, fill=pink!10!blue!80!red!10, text width=3cm, text=black]]
                    [AG, color=blue!40, fill=blue!20, text width=1.2em, text=black [HEF \cite{yang2024enhancingerg}; Bhaskar et al. \cite{bhaskar-etal-2023-prompted}; EmpCRL \cite{cai-etal-2024-empcrl}; Sibyl \cite{wang-etal-2025-sibyl}; TOOL-ED \cite{cao-etal-2025-tool}, color=pink!40!blue!80!red!40, fill=pink!10!blue!80!red!10, text width=3cm, text=black]]
                ]
                [Knowledge Distillation, color=pink, fill=pink!60, text width=6em, text=black
                    [AU, color=blue!40, fill=blue!20, text width=1.2em, text=black [SCRAP \cite{kim-etal-2024-self-consistent}; Distill \cite{zhang2024distillingfinegrained}; Cao et al. \cite{CaoImprovingESC}, color=pink!40!blue!80!red!40, fill=pink!10!blue!80!red!10, text width=3cm, text=black]]
                    [AG, color=blue!40, fill=blue!20, text width=1.2em, text=black [Sibyl \cite{wang-etal-2025-sibyl}, color=pink!40!blue!80!red!40, fill=pink!10!blue!80!red!10, text width=3cm, text=black]]
                ] 
            ]
            [\textbf{Instance-level Augmentation} (\textsection~\ref{INSTANCE}), color=pink, fill=pink!60, text width=6em, text=black
                [Synthetic Data Augmentation, color=pink, fill=pink!60, text width=6em, text=black
                    [AU, color=blue!40, fill=blue!20, text width=1.2em, text=black [DS$^2$-ABSA \cite{xu2024ds2absadualstream}; BERT + CADA\cite{XuXQWT25}; IDG \cite{zhong2024iterative}, color=pink!40!blue!80!red!40, fill=pink!10!blue!80!red!10, text width=3cm, text=black]]
                    [AG, color=blue!40, fill=blue!20, text width=1.2em, text=black [ChatPal \cite{zheng-etal-2024-self}; STICKERCONV\cite{zhang-etal-2024-stickerconv}; SoulChat \cite{chen-etal-2023-soulchat}, color=pink!40!blue!80!red!40, fill=pink!10!blue!80!red!10, text width=3cm, text=black]]
                ]
                [Auxiliary Task Augmentation, color=pink, fill=pink!60, text width=6em, text=black
                    [AU, color=blue!40, fill=blue!20, text width=1.2em, text=black [RVISA \cite{rvisa}; InstructERC \cite{lei2024instructerc}; MT-ISA \cite{lai2024mtisa}; CKERC \cite{fu2024ckercjointlarge}; ECR-Chain \cite{ijcai2024ECRChain}, color=pink!40!blue!80!red!40, fill=pink!10!blue!80!red!10, text width=3cm, text=black]]
                ]
            ]
            [\textbf{Modal-level Augmentation} (\textsection~\ref{MODAL}), color=pink, fill=pink!60, text width=6em, text=black
                [Multimodal Affective Signal Encoding, color=pink, fill=pink!60, text width=6em, text=black
                    [AU, color=blue!40, fill=blue!20, text width=1.2em, text=black [EmoLLM \cite{yang2024emollm}; MERITS-L \cite{dutta2025llmsupervisedpretrainingmultimodal}, color=pink!40!blue!80!red!40, fill=pink!10!blue!80!red!10, text width=3cm, text=black]]
                    [AG, color=blue!40, fill=blue!20, text width=1.2em, text=black [STICKERCONV \cite{zhang-etal-2024-stickerconv}, color=pink!40!blue!80!red!40, fill=pink!10!blue!80!red!10, text width=3cm, text=black]]
                ]
                [Multimodal Affective Cues Decoding, color=pink, fill=pink!60, text width=6em, text=black
                    [AU, color=blue!40, fill=blue!20, text width=1.2em, text=black [DialogueLLM \cite{zhang2024dialoguellm}; Feng et al. \cite{feng-etal-2024-affect}; WisdoM\cite{mm/Wang0S00T24}, color=pink!40!blue!80!red!40, fill=pink!10!blue!80!red!10, text width=3cm, text=black]]
                    [AG, color=blue!40, fill=blue!20, text width=1.2em, text=black [SMES \cite{chu2024multimodalemo}; ELMD \cite{Liu2024SpeakFromHeart}, color=pink!40!blue!80!red!40, fill=pink!10!blue!80!red!10, text width=3cm, text=black]]
                ]
            ]
        ]
        [\textbf{Autonomous Collaboration}, color=pink, fill=pink!60, text width=6em, text=black
            [\textbf{Multi-Agent Systems} (\textsection~\ref{MAS}), color=pink, fill=pink!60, text width=6em, text=black
                [Consensus-driven Interaction, color=pink, fill=pink!60, text width=6em, text=black
                    [AU, color=blue!40, fill=blue!20, text width=1.2em, text=black [Negotiation \cite{sun2023sentimentanalysisllmnegotiations}; HAD \cite{FrankAgentforSA}, color=pink!40!blue!80!red!40, fill=pink!10!blue!80!red!10, text width=3cm, text=black]]
                ]
                [Scenario-immersive Simulation, color=pink, fill=pink!60, text width=6em, text=black
                    [AG, color=blue!40, fill=blue!20, text width=1.2em, text=black [STICKERCONV \cite{zhang-etal-2024-stickerconv}; Cooper \cite{aaai/ChengLWLOLW024}; VLESA \cite{li-etal-2024-helpful}, color=pink!40!blue!80!red!40, fill=pink!10!blue!80!red!10, text width=3cm, text=black]]
                ]
            ]   
        ]
    ],
    ]
]
\end{forest}
}
\caption{Taxonomy of LLM-based collaboration strategies with related references, organized separately for the affective understanding (AU) and affective generation (AG) domains.}
   \label{collaboration modes}
\end{figure*}

%% file: src/experiment.tex
\section{Experiment}
\label{experiments}
In this section, we systematically evaluate the efficacy of LLM-based collaboration strategies across three representative tasks in AC: aspect-based sentiment analysis (ABSA) and emotion recognition in conversation (ERC) within the AU domain, and emotion support conversation (ESC) within the AG domain. Our objective is to provide a comprehensive comparison of the advantages and limitations inherent in various collaboration strategies. The experimental results and analyses are detailed in the subsequent sections.

\subsection{LLM-based Collaboration in ABSA}
\subsubsection{Experiment Setup}
\paragraph{Datasets}
We utilize the Restaurant and Laptop datasets from SemEval-2014 \cite{Pontiki_semeval_2014} to assess the performance of ABSA tasks. These datasets comprise text reviews from the laptop and restaurant domains, with the goal of identifying sentiment polarity concerning specific aspects or target terms mentioned within review sentences, categorizing them as positive, negative, or neutral.

\paragraph{Metrics}
The evaluation metrics include the \emph{macro-F1 score} to measure task performance. Additionally, we assess the \emph{API calls} made during training or inference to gauge commercial costs associated with LLM usage. The \emph{train model size} reflects the parameters required for updates, indicating computational costs alongside the \emph{assist model size}.

\paragraph{Compared Methods}
\begin{itemize}
    \item \textbf{Baselines}
    The baseline methods include \emph{Zero-shot} approaches, which directly prompt the backbone model for inference, and \emph{Supervised Fine-tuning} (SFT), which involves fine-tuning the backbone inference model. Models in the \emph{Zero-shot} setting include GPT3+CoT, GPT-4o-mini, and GPT-4o, while those in \emph{SFT} comprise T5-Base \cite{Raffel2019ExploringTL}, Flan-T5 (base and XXL size) \cite{flant5}, Llama-2-7B \cite{touvron2023llama2openfoundation}, and Mixtral-8x7B \cite{jiang2024mixtralexperts}.
    
    \item \textbf{Token-level Augmentation}
    Token-level augmentation involves two methods: MLKF \cite{cikm/LiuZLZC24}, which applied a knowledge augmentation strategy for collaboration by integrating heterogeneous knowledge from multiple LLMs for cognitive reasoning in ABSA, and Distill \cite{zhang2024distillingfinegrained}, which employed a knowledge distillation strategy that generates knowledge data based on analysis and rewriting prompts from the teacher model, subsequently distilling the knowledge to the student model for ABSA.
    \item \textbf{Instance-level Augmentation}
    Instance-level augmentation encompasses both synthetic data augmentation and auxiliary task augmentation, with two evaluated methods in each category. 
    In synthetic data augmentation, BERT + CADA \cite{XuXQWT25} utilized LLM for context- and aspect-aware synthetic data augmentation, adopting contrastive learning on the BERT \cite{Devlin_Chang_Lee_Toutanova_2019} model. BERT-SPC + IDG \cite{zhong2024iterative} proposed a three-stage framework to iteratively augment data from LLMs, applying the augmented data to existing methods such as BERT-SPC \cite{BERT-SPC}, which conducted sentence pair classification using the BERT model. 
    In auxiliary task augmentation, RVISA \cite{rvisa} was augmented with explanation and verification tasks using LLMs for reasoning in ABSA, fine-tuning with Flan-T5 \cite{flant5}. MT-ISA \cite{lai2024mtisa} applied LLMs for auxiliary task construction with a self-refinement strategy from the perspectives of sentiment elements, fine-tuning Flan-T5 \cite{flant5} in a multi-task learning paradigm with automatic weight learning.
    \item \textbf{Multi-agent Systems}
    The multi-agent system employs the Negotiation \cite{sun2023sentimentanalysisllmnegotiations} method, implemented with various combinations of LLM agents as discriminator and generator for consensus-driven interaction. Role flipping is conducted for majority voting when LLMs with initial roles fail to reach consensus after negotiation.
\end{itemize}
Since the evaluated dataset is text-only, modal-level augmentation is omitted in the ABSA evaluation and will be compared in the subsequent tasks.

\begin{table*}[h!t] 
  \caption{The experiment results on ABSA datasets with different LLM-based collaboration strategies, including Token-level Augmentation, Instance-level Augmentation, and Multi-agent Systems. The bold text indicates the best result, while the underlined text represents the second-best result.}
  \label{tab:ABSA evaluation}
  \centering
  \resizebox{\linewidth}{!}{
  \begin{tabular}{llllllllcc}
    \toprule[1.2pt]
    
        \multicolumn{2}{l}{\multirow{2}{*}{\hfil \textbf{Collaboration Strategies}}} & \multirow{2}{*}{\hfil \textbf{Methods}} & \multirow{2}{*}{\hfil \textbf{Assist Model}}  & \multirow{2}{*}{\hfil \textbf{Assist Model Size}} & \multirow{2}{*}{\hfil \textbf{Inference Model}} & \multirow{2}{*}{\hfil \textbf{Train Model Size}} & \multirow{2}{*}{\hfil \textbf{API Calls}} & \multicolumn{2}{c}{\textbf{F1 Score} (marco)}\\
        % \cline{9-10}
        \cmidrule{9-10}
         \multicolumn{2}{l}{}&&&&&&& \textbf{Rest14} & \textbf{Lap14}\\
        % \hline
        \midrule
        \multirow{8}{*}{\hfil Baselines}&\multirow{8}{*}{\hfil -} & \multirow{3}{*}{\hfil Zero-shot} & \multirow{3}{*}{\hfil-}&\multirow{3}{*}{\hfil-}&GPT3+CoT& Training-free & \multirow{3}{*}{\hfil-}& 68.32&66.65\\
       &&&&&GPT-4o-mini&Training-free&& 72.52&73.83\\
       &&&&&GPT-4o&Training-free&& 73.82&74.24\\
       \cmidrule{3-10}
       & &\multirow{5}{*}{\hfil SFT}  &\multirow{5}{*}{\hfil -}& \multirow{5}{*}{\hfil -} &T5-Base & 220M & \multirow{5}{*}{\hfil -}& 67.33	& 60.78\\
        &&&&&Flan-T5-Base&220M&& 79.78 & 77.93\\
        &&&&&Llama-2-7B&7B&& 70.72 &63.39\\
        &&&&&Flan-T5-XXL&11B&& 83.68&77.69\\
       &&&& &Mixtral-8x7B&46.7B&& 72.59&64.40\\
       % \cline{3-10}
       % \hline
       \midrule
        \multirow{4}{*}{\hfil \begin{tabular}[c]{@{}l@{}} Token-level \\ Augmentation\end{tabular}}&Knowledge Augmentation  & MLKF \cite{cikm/LiuZLZC24}&\begin{tabular}[c]{@{}l@{}}Vicuna-13B \\+ WizardLM-13B \end{tabular}&13B*2&Vicuna-13B  & Training-free & - & 73.97&71.19\\
        % \cline{2-10}
        \cmidrule{2-10}
        &\multirow{3}{*}{\hfil Knowledge Distillation}  & Distill \cite{zhang2024distillingfinegrained}&Llama-2-7B &7B&T5-Base  & 220M & \textasciitilde 2M & 69.96&63.93\\
        && Distill \cite{zhang2024distillingfinegrained}&Mixtral-8x7B &46.7B &T5-Base & 220M & \textasciitilde 2M & 71.21&65.04\\
        && Distill \cite{zhang2024distillingfinegrained}&GPT-3.5-turbo &API &T5-Base & 220M & \textasciitilde 200K & 69.45&62.96\\
        % \hline
        \midrule
        \multirow{6}{*}{\hfil \begin{tabular}[c]{@{}l@{}} Instance-level \\ Augmentation\end{tabular}}&\multirow{2}{*}{\hfil \begin{tabular}[c]{@{}l@{}} Synthetic Data \\Augmentation\end{tabular}}  & BERT + CADA \cite{XuXQWT25} &GPT-3.5-turbo &API&BERT  & 110M & \textasciitilde 18K & 81.26&77.71\\
        & & BERT-SPC + IDG \cite{zhong2024iterative}&GPT-3.5-turbo &API &BERT  & 110M & \textasciitilde 30K & 77.13&78.45\\
        % \cline{2-10}
        \cmidrule{2-10}
        &\multirow{4}{*}{\hfil \begin{tabular}[c]{@{}l@{}} Auxiliary Task \\Augmentation\end{tabular}} & RVISA \cite{rvisa}& GPT-3.5-turbo &API &Flan-T5-XXL & 11B & \textasciitilde 12K & 86.85&\textbf{86.85}\\
        && RVISA \cite{rvisa}& GPT-4o &API &Flan-T5-XXL & 11B & \textasciitilde 12K & \underline{87.55}&\underline{84.48}\\
        && MT-ISA \cite{lai2024mtisa}& GPT-4o-mini  &API &Flan-T5-Base & 220M & \textasciitilde 24K & 82.45&79.86\\
        && MT-ISA \cite{lai2024mtisa}& GPT-4o-mini  &API &Flan-T5-XXL  & 11B & \textasciitilde 24K & \textbf{88.96}&83.86\\
        % \hline
        \midrule
        \multirow{2}{*}{\hfil\begin{tabular}[c]{@{}l@{}} Multi-Agent \\ Systems \end{tabular}}& \multirow{2}{*}{\hfil\begin{tabular}[c]{@{}l@{}} Consensus-driven \\Interaction \end{tabular}}&  \multirow{2}{*}{\hfil Negotiation \cite{sun2023sentimentanalysisllmnegotiations}} &\multirow{2}{*}{\hfil -}&\multirow{2}{*}{\hfil -}&GPT-4o + DeepSeek-R1& Training-free & \textasciitilde 6K &81.06 &81.92\\
        % \cmidrule{3-10}
        && &&&DeepSeek-V3 + DeepSeek-R1& Training-free & \textasciitilde 6K &81.89 &76.40\\

    \bottomrule[1.2pt]
  \end{tabular}
  }
\end{table*}
\subsubsection{Experimental Results}
Table \ref{tab:ABSA evaluation} presents the experimental results across different LLM-based collaboration strategies. Instance-level augmentation consistently demonstrates superior performance compared to other collaboration strategies in ABSA. Within token-level augmentation, MLKF \cite{cikm/LiuZLZC24} enhances the more robust backbone model Vicuna-13B with domain knowledge for inference, outperforming the Distill \cite{zhang2024distillingfinegrained} method, which fine-tunes a smaller student model despite leveraging additional datasets. This suggests that LLMs, empowered with remarkable reasoning and generation capabilities, hold potential for effectively addressing ABSA tasks through in-context learning principles.

When the Distill \cite{zhang2024distillingfinegrained} method increased augmented knowledge data from 200k to 2M, assistance from less powerful teacher models like Llama-2-7B and Mixtral-8x7B yielded better performance than assistance from the more powerful GPT-3.5-turbo model. This underscores the importance of balancing assist model size with the quality and diversity of augmented data in structured collaboration. Similarly, in instance-level augmentation, both BERT + CADA \cite{XuXQWT25} and BERT-SPC + IDG \cite{zhong2024iterative} employed the same assist and inference models but differed in synthetic data augmentation strategies applied to the assist model, with performance primarily influenced by augmented data quality. RVISA and MT-ISA, utilizing more powerful assist and inference models while carefully designing verification mechanisms for synthetic generation, exhibit superior performance compared to methods in synthetic data augmentation.

For autonomous collaboration using multi-agent systems, the Negotiation \cite{sun2023sentimentanalysisllmnegotiations} method surpasses token-level augmentation and some instance-level augmentation methods employing smaller inference models. This highlights the potential of autonomous collaboration with existing off-the-shelf LLMs in addressing ABSA tasks, avoiding heavy computational costs associated with model training but incurring API call costs for real-time data inference. In contrast, both token-level and instance-level strategies primarily introduce one-time API calls for training data.

\subsection{LLM-based Collaboration in ERC}
\subsubsection{Experiment Setup}
\paragraph{Datasets}
We evaluated three benchmarks for ERC: IEMOCAP \cite{IEMOCAP}, EmoryNLP \cite{EmoryNLP}, and MELD \cite{MELD}. IEMOCAP \cite{IEMOCAP} is a widely used multimodal dataset that annotates $6$ types of emotions: happy, sad, neutral, angry, excited, and frustrated. It contains more than $7,433$ utterances and $151$ dialogues. EmoryNLP \cite{EmoryNLP} is collected from the TV series \emph{Friends} and comprises utterances categorized into $7$ distinct emotional classes: neutral, joyful, peaceful, powerful, scared, mad, and sad.. It includes $97$ episodes, $897$ scenes, and $12606$ utterances. MELD \cite{MELD} is also a multimodal dataset derived from the TV series \emph{Friends} and covers $7$ different emotional categories: anger, disgust, fear, happiness, neutrality, sadness, and surprise. It contains raw dialogue transcripts, audio, video, and emotional annotations with a total of $13,708$ utterances.

\paragraph{Metrics}
The task performance evaluation metric uses the \emph{weighted-F1 score}. The \emph{assist model size} and \emph{train model size} indicate the computational cost required for each method.

\begin{itemize}
    \item \textbf{Baselines}
     The baseline methods include \emph{Zero-shot} approaches, which directly prompt the backbone model for inference, and \emph{LoRA Fine-tuning}, which applies LoRA techniques \cite{Hu_Shen_Wallis_Allen-Zhu_Li_Wang_Chen_2021} for parameter-efficient fine-tuning on the backbone inference model. Models in the \emph{Zero-shot} setting includes ChatGLM \cite{glm2024chatglmfamilylargelanguage} and ChatGPT3.5 \cite{CHATGPT}, while those in \emph{SFT} included Llama-7B \cite{touvron2023llamaopenefficientfoundation} and Llama-2-7B \cite{touvron2023llama2openfoundation}.
    \item \textbf{Token-level Augmentation}
    InstructERC \cite{lei2024instructerc} applied the retrieval-augmented demonstration strategy, using Sentence-BERT \cite{sentence-bert} for embedding similarity check to retrieve augmented demonstration conducive to instruction tuning on inference model.
    \item \textbf{Instance-level Augmentation}
     CKERC \cite{fu2024ckercjointlarge} used additional LLMs to augment the auxiliary speaker commonsense identification task for first-stage training and then conducted LoRA fine-tuning on the InstructERC \cite{lei2024instructerc} model for the second-stage training that determined the final emotion.
    \item \textbf{Modal-level Augmentation}
    MERITS-L \cite{dutta2025llmsupervisedpretrainingmultimodal} applied a multimodal affective signal encoding strategy to align the audio signal with the text signal by pre-training on LLM-annotated transcripts and then joint training with an additional audio encoder. In contrast, DialogueLLM \cite{zhang2024dialoguellm} decoded video signals into affective cues in text format from multimodal LLM, ERNIE Bot, as augmented data for inference model training.
    \item \textbf{Multi-Agent Systems}
    The multi-agent system employs the Negotiation \cite{sun2023sentimentanalysisllmnegotiations} method, implemented with various combinations of LLM agents as discriminator and generator for consensus-driven interaction. Role flipping is conducted for majority voting when LLMs with initial roles fail to reach consensus after negotiation. In the implementation, the input prompt setting is from InstructERC \cite{lei2024instructerc}, with role-specific instructions from Negotiation \cite{sun2023sentimentanalysisllmnegotiations}.
\end{itemize}

\begin{table*}[h!t]
    \caption{The experiment results on the ERC task across three datasets, with different LLM-based collaboration strategies, including Token-level Augmentation, Instance-level Augmentation, Modal-level Augmentation, and Multi-agent Systems. The bold text indicates the best result, while the underlined text represents the second-best result.}
    \label{tab:ERC evaluation}
    \centering
    \resizebox{\linewidth}{!}{
    % \begin{tabular}{|l|l|l|l|l|l|l|l|l|l|l|}
    \begin{tabular}{lllllllllll}
         \toprule[1.2pt]
         \multicolumn{2}{l}{\multirow{2}{*}{\hfil \textbf{Collaboration Strategies}}} & \multirow{2}{*}{\hfil \textbf{Methods}} & \multirow{2}{*}{\hfil \textbf{Assist Model}} &  \multirow{2}{*}{\hfil  \textbf{Assist Model Size}} &\multirow{2}{*}{\hfil \textbf{Inference Model}} & \multirow{2}{*}{\hfil \textbf{Train Model Size}} & \multirow{2}{*}{\hfil \textbf{Modality}} & \multicolumn{3}{c}{\textbf{F1 Score} (weighted)}\\
         % \cline{9-11}
         \cmidrule{9-11}
          \multicolumn{2}{l}{}  & &  &  & &  & & \textbf{IEMOCAP} & \textbf{EmoryNLP} & \textbf{MELD} \\ \hline
        \multirow{4}{*}{\hfil Baseline} &\multirow{4}{*}{\hfil-}& \multirow{2}{*}{\hfil Zero-shot }&  \multirow{2}{*}{-} &   \multirow{2}{*}{-} &  ChatGLM & Training-free&Text & 38.60&19.60&38.80 \\ 

        &  & &   & & ChatGPT3.5 & Training-free&Text & 53.38&37.00&65.07 \\  
        \cmidrule{3-11}
        & &\multirow{2}{*}{\hfil LoRA Fine-tuning }& \multirow{2}{*}{-}  & \multirow{2}{*}{\hfil-} & LLaMA-7B &  7B (LoRA) →12.5M& Text & 55.81&37.98&66.15 \\ 
        % \cline{6-11}
        % \cmidrule{6-11}
        &  & &   & & LLaMA-2-7B&  7B (LoRA) →12.5M& Text & 55.96 &38.21&65.84 \\ 
        % \hline
        \midrule
         \begin{tabular}[c]{@{}l@{}}Token-level \\Augmentation\end{tabular}&  \begin{tabular}[c]{@{}l@{}} Retrieval Augmented \\ Demonstration \end{tabular} & InstructERC \cite{lei2024instructerc} &Sentence-BERT& 110M& LLaMA-2-7B&  7B (LoRA) →12.5M &Text &71.39&\underline{41.37}&69.15\\
        % \cline{6-11}
        % \cmidrule{6-11}
        
        % \hline
        \midrule
         \begin{tabular}[c]{@{}l@{}} Instance-level\\ Augmentation\end{tabular}  &  \begin{tabular}[c]{@{}l@{}} Auxiliary Task \\ Augmentation \end{tabular}  & CKERC \cite{fu2024ckercjointlarge}& LLaMA-2-7B-chat&7B  & LLaMA-2-7B & 7B (LoRA) & Text &\underline{72.40}&\textbf{42.08}&\underline{69.27} \\ 
        % \hline
        \midrule
        \multirow{4}{*}{\begin{tabular}[c]{@{}l@{}}Modal-level \\Augmentation\end{tabular}} & \begin{tabular}[c]{@{}l@{}}Multimodal Affective \\Signal Encoding\end{tabular} & MERITS-L \cite{dutta2025llmsupervisedpretrainingmultimodal}& \begin{tabular}[c]{@{}l@{}}GPT-3.5-turbo \\ + Whisper-large-v3\end{tabular}  & API + 1.54B& RoBERTa-large  & 355M & Audio+Text & \textbf{86.48}	&-&	66.02 \\ 
        % \cline{2-11}
        \cmidrule{2-11}
         & \begin{tabular}[c]{@{}l@{}}Multimodal Affective \\Cues Decoding\end{tabular} &  DialogueLLM \cite{zhang2024dialoguellm}& ERNIE Bot & 260B & LLaMA-2-7B   &  7B (LoRA) →4.2M& Video+Text & 69.93	&40.05&\textbf{71.90} \\ 
         % \hline
         \midrule
        \multirow{2}{*}{\hfil\begin{tabular}[c]{@{}l@{}} Multi-Agent \\ Systems \end{tabular}}& \multirow{2}{*}{\hfil\begin{tabular}[c]{@{}l@{}} Consensus-driven \\Interaction \end{tabular}}&   \multirow{2}{*}{\hfil Negotiation \cite{sun2023sentimentanalysisllmnegotiations}} &\multirow{2}{*}{\hfil -}&\multirow{2}{*}{\hfil -}&GPT-4o + DeepSeek-R1& Training-free & Text &49.58&37.13&62.54\\
        % \cmidrule{3-11}
        &&&&&DeepSeek-V3 + DeepSeek-R1& Training-free & Text &51.68&38.94&62.07\\
        % \hline
        \bottomrule[1.2pt]
    \end{tabular}
    }
\end{table*}
\subsubsection{Experimental Results}
The experimental results are presented in Table \ref{tab:ERC evaluation}. Unlike ABSA, LLMs (e.g., ChatGPT \cite{CHATGPT}) without task-specific training lag behind LLaMA models that have been trained using LoRA \cite{Hu_Shen_Wallis_Allen-Zhu_Li_Wang_Chen_2021} techniques. Even when employing an autonomous collaboration strategy using Negotiation \cite{sun2023sentimentanalysisllmnegotiations} settings with the most advanced LLMs, including GPT-4o and Deepseek, the performance remains inferior to supervised methods. This underscores the complexity of the ERC task, which demands context understanding in lengthy conversations where emotions change dynamically and frequently due to real-time interactions. Consequently, LLMs struggle to accurately capture emotional shifts by relying solely on pre-trained knowledge and task instructions.

Structured collaboration with various levels of augmentation demonstrates significant performance improvements compared to baseline methods. The CKERC \cite{fu2024ckercjointlarge} method in instance-level augmentation introduced an auxiliary task with additional speaker commonsense knowledge from the LLaMA model, surpassing the InstructERC \cite{lei2024instructerc} method in token-level augmentation, which collaborated with a smaller assist model for retrieving similar demonstrations in task-specific prompts only. However, nuanced emotions can be conveyed through tones, facial expressions, and body language. Modal-level augmentation aids in capturing more affective cues by fully utilizing data across different modalities, as evidenced by IEMOCAP \cite{IEMOCAP} and MELD \cite{MELD}, which involve multimodal data. The primary challenge lies in aligning multimodal data with recognized emotions. Both MERITS-L \cite{dutta2025llmsupervisedpretrainingmultimodal}, which applied a multimodal affective signal encoding strategy to fuse multimodal signals with modal-specific encoders, and DialogueLLM \cite{zhang2024dialoguellm}, which applied a multimodal affective cues decoding strategy to uniformly handle multimodal data in text format by transforming different modal data from multimodal LLMs, exhibit performance increases compared to baseline methods by enhancing modal-level feature perception. There remains room for improved alignment and task enhancement in some datasets compared to token-level and instance-level augmentation, which focus solely on text modality.

\subsection{LLM-based Collaboration in ESC}
\subsubsection{Experiment Setup}

\paragraph{Datasets}
The evaluation of the ESC task utilizes the ESConv dataset \cite{Liu2021TowardsES}. ESConv comprises approximately $1,300$ multi-turn dialogues between help seekers experiencing emotional distress and professional supporters, providing a rich context for assessing the effectiveness of various collaboration strategies.
\paragraph{Metrics}
The evaluation employs several automatic metrics. BLEU-1/2/3/4 (\textbf{BL-1/2/3/4}) \cite{Bleu}, evaluates the precision of n-grams, providing insight into how well the generated text matches the reference text; ROUGE-L (\textbf{RG-L}) \cite{lin-2004-rouge}measures the recall of the longest common subsequences, indicating the extent to which the generated text captures the essential elements of the reference; METEOR (\textbf{MET}) \cite{METEOR}takes into account stem matches and synonyms, offering a more nuanced evaluation of semantic similarity; Distinct-1/2/3 (\textbf{Dist-1/2/3}) \cite{DIST} calculates the proportion of unique n-grams, reflecting the diversity and originality of the generated text. Additionally, the \emph{assist model size} and \emph{train model size} are considered to indicate the computational cost associated with each method.
\paragraph{Compared Methods}
\begin{itemize}
    \item \textbf{Baselines}
    The baseline methods included \emph{Zero-shot} approaches, which directly prompted the backbone model for inference, and \emph{Supervised Fine-tuning} (SFT), which involved fine-tuning the backbone inference model. Models in the \emph{Zero-shot} setting included GPT-3.5, GPT3.5+CoT, and GPT-4o, while those in \emph{SFT} comprised Llama-3.1.
    \item \textbf{Token-level Augmentation}
    Sibyl \cite{wang-etal-2025-sibyl} applied both knowledge augmentation and knowledge distillation strategies for the ESC task. It first distilled a student model, LLaMA-7B, from GPT-4o to decide visionary commonsense knowledge, and then augmented the knowledge from the well-trained student model for subsequent model training on emotional support dialogues.
    \item \textbf{Multi-agent Systems}
    Two methods are evaluated with the scenario-immersive simulation strategy. Cooper \cite{aaai/ChengLWLOLW024} coordinated multiple specialized agents to facilitate achieving complex dialogue goals, with each agent dynamically progressing towards the final goal aspect based on real-time interaction states. VLESA \cite{li-etal-2024-helpful} aligned agents with a helpfulness scorer using a BERT \cite{Devlin_Chang_Lee_Toutanova_2019} model and developed optimal relevance learning to refine agent actions constrained by the scorer.
\end{itemize}
Since the evaluated dataset is text-only, modal-level augmentation is omitted for comparison. In the case of instance-level augmentation, previous studies on synthetic data augmentation \cite{chen-etal-2023-soulchat,zheng-etal-2024-self} typically involve the creation of proprietary datasets for evaluation. To maintain consistency and fairness in our comparison with ESConv, these methods are not employed here.

\begin{table*}[h!t]
    \caption{The experiment results for the ESC task on ESConv \cite{Liu2021TowardsES} dataset, with different LLM-based collaboration strategies, including Token-level Augmentation and Multi-agent Systems. The bold text indicates the best result, while the underlined text represents the second-best result.}
    \label{tab:ESC evaluation}
    \centering
    \resizebox{\linewidth}{!}{
    % \begin{tabular}{|l|l|l|l|l|l|l|c|c|c|c|c|c|c|c|c|}
    \begin{tabular}{lllllllccccccccc}
        \toprule[1.2pt]
         \multicolumn{2}{l}{\multirow{2}{*}{\hfil \textbf{Collaboration Strategies}}} & \multirow{2}{*}{\hfil \textbf{Methods}} & \multirow{2}{*}{\hfil \textbf{Assist Model}} &  \multirow{2}{*}{\hfil  \textbf{Assist Model Size}} &\multirow{2}{*}{\hfil \textbf{Inference Model}} & \multirow{2}{*}{\hfil \textbf{Train Model Size}} & \multicolumn{6}{c}{\textbf{NLG Metrics}} & \multicolumn{3}{c}{\textbf{Diversity}}\\
         \cmidrule(lr){8-13} \cmidrule(lr){14-16}
          \multicolumn{2}{l}{}&&&&&& \textbf{BL-1}&\textbf{BL-2}&\textbf{BL-3}&\textbf{BL-4}&\textbf{RG-L}&\textbf{MET}&\textbf{Dist-1}&\textbf{Dist-2}&\textbf{Dist-3}\\
         \midrule
         \multirow{4}{*}{\hfil Baseline} &\multirow{4}{*}{\hfil-} & \multirow{3}{*}{\hfil Zero-shot} & \multirow{3}{*}{-}  & \multirow{3}{*}{\hfil-} & GPT-3.5&Training-free&17.16&5.04&-&1.02&15.44&9.12&4.50&25.53&47.72\\
         \cmidrule{6-16}
         &&&&&GPT-3.5+CoT&Training-free&15.86&4.66&-&0.94&14.42&9.36&4.29&24.61&47.62\\
         \cmidrule{6-16}
         &&&&&GPT-4o&Training-free&-&5.06&2.01&0.93&14.86&8.50&6.43&31.39&56.38\\
         \cmidrule{3-16}
         && SFT & -  & - & Llama 3.1 & 8B & - & 6.75&2.92&	1.41&15.62&9.12&6.24&\underline{40.34}&\underline{75.60}\\
         \midrule
         \multirow{3}{*}{\hfil \begin{tabular}[c]{@{}l@{}} Token-level \\Augmentation \end{tabular}} &\multirow{3}{*}{\hfil \begin{tabular}[c]{@{}l@{}}Knowledge Distillation\\ + Augmentation\end{tabular}} & \multirow{3}{*}{\hfil Sibyl \cite{wang-etal-2025-sibyl}} & \multirow{3}{*}{\hfil \begin{tabular}[c]{@{}l@{}}GPT-4o \\+ LLaMA-7B \end{tabular}}&\multirow{3}{*}{API + 7B }& GPT-4o& 7B (LoRA)→ 4.2M (Assist)&-&5.19&2.21&1.10&15.20&\underline{9.65}&\underline{6.52}&32.09&56.72\\
         \cmidrule{6-16}
         &&&& & Llama 3.1& \begin{tabular}[c]{@{}l@{}} 7B (LoRA)→ 4.2M (Assist)\\+ 8B (Inference)\end{tabular} & -&6.97&3.04&1.52&16.23&8.53&\textbf{6.84}&\textbf{41.59}&\textbf{76.41}\\
         \midrule
         \multirow{7}{*}{\hfil \begin{tabular}[c]{@{}l@{}} Multi-Agent\\ Systems \end{tabular}} &\multirow{7}{*}{\hfil \begin{tabular}[c]{@{}l@{}} Scenario-immersive \\ Simulation\end{tabular}} & \multirow{3}{*}{\hfil Cooper \cite{aaai/ChengLWLOLW024}} & \multirow{3}{*}{\hfil \begin{tabular}[c]{@{}l@{}}GPT-3.5-turbo*3 \\ + MPNet\end{tabular} }  & \multirow{3}{*}{\hfil API*3 + 110M} & GPT-3.5-turbo & A few millions (Assist) &17.62&5.42&-&1.11&15.86&9.36&5.22&29.45&54.40\\
         \cmidrule{6-16}
         &&& &  & BART & \begin{tabular}[c]{@{}l@{}} A few millions (Assist)\\+ 139M (Inference) \end{tabular}  &\underline{22.76}&\underline{9.54}&-&\underline{3.11}&\textbf{20.18}&9.22&5.02&24.22&43.55\\
         \cmidrule{3-16}
         &&\multirow{4}{*}{\hfil VLESA \cite{li-etal-2024-helpful}}&\multirow{4}{*}{\hfil \begin{tabular}[c]{@{}l@{}}LlaMA-2-7b-chat  \\ + Bert-Base\end{tabular}}  & \multirow{4}{*}{\hfil 7B + 110M } & BlenderBot-small & \begin{tabular}[c]{@{}l@{}} 110M (Assist) \\+ 90M (Inference) \end{tabular}  &20.84&8.78&\underline{4.55}&2.67&18.09&9.00&-&-&-\\
         \cmidrule{6-16}
         &&&  & & BART & \begin{tabular}[c]{@{}l@{}} 110M (Assist)\\+ 139M (Inference) \end{tabular}  &\textbf{23.53}&\textbf{9.97}&\textbf{5.30}&\textbf{3.17}&\underline{19.96}&\textbf{9.74}&-&-&-\\
         \bottomrule[1.2pt]
    \end{tabular}
    }
\end{table*}

\subsubsection{Experimental Results}
The experimental results are presented in Table \ref{tab:ESC evaluation}. The Llama 3.1 model, fine-tuned with domain-specific data, surpasses GPT-series models operating in a zero-shot capacity. Similarly, Sibyl \cite{wang-etal-2025-sibyl}, trained on Llama 3.1 and augmented with additional visionary commonsense knowledge from the distilled model, exhibits superior performance on automatic metrics compared to direct inference with GPT-4o that without task-specific training. This observation aligns with trends seen in ERC tasks, where models that undergo task-specific training and augmentation consistently outperform off-the-shelf LLMs that rely solely on pre-trained, generalized capabilities. This is attributed to the inherent complexity of both ESC and ERC tasks, which demand sophisticated reasoning and generation over extended contexts. Such tasks benefit significantly from models that are specifically tailored or augmented to handle the intricacies of long-context interactions and emotional nuances.

In autonomous collaboration with multi-agent systems, both Cooper and VLESA utilized LLMs as agents for scenario-immersive simulation. With specialized models trained to align with agent interactions, they demonstrated the highest performance in the ESC task, which requires both affective understanding and generation capabilities, while incurring less computational cost on model training. This underscores the potential of LLM-based autonomous collaboration to perform satisfactorily under certain constraints, which can be derived from well-trained specialized models. However, real-time agents and API calls are necessary for autonomous interaction, potentially leading to higher service costs when increasing the data volume or interaction duration.

%% file: src/discussion.tex
\section{Discussion}
\label{discuss}
\subsection{Highlight of Findings}
\subsubsection{Aspect-based Sentiment Analysis (ABSA)} 
The experimental results demonstrate that instance-level augmentation strategies significantly outperform token-level augmentation and multi-agent systems (MAS) in ABSA. For such tasks requiring fine-grained sentiment understanding, high-quality and diverse augmented data contribute to improving model performance. Additionally, balancing the size of assist models with the quality of augmented data is crucial; even smaller models aided by rich augmented data can achieve competitive results, highlighting the importance of data quality over sheer model size. While MAS can surpass most of the supervised methods trained with small models in ABSA and serve as an effective solution under limited computational resources, they may introduce higher API calls in real-time scenarios when increasing data volume.
\subsubsection{Emotion Recognition in Conversation (ERC)} 
ERC presents unique challenges due to the necessity of understanding long conversational contexts and dynamic emotional shifts. Modal-level augmentation can effectively capture nuanced affective cues across different modalities, such as tone and facial expressions, enriching emotional understanding through multimodal information. While other structured collaboration strategies, including token-level and instance-level augmentation with task-specific training on the inference model, can produce great enhancement in model performance compared to zero-shot or LoRA fine-tuning on a single model, even surpassing the MAS setting with consensus-driven interaction. These observations highlight the inherent difficulty for off-the-shelf LLMs to handle long-context and nuanced emotion categories without extensive training, emphasizing the greater complexity of ERC tasks relative to the aforementioned ABSA.
\subsubsection{Emotion Support Conversation (ESC)}  
ESC demands robust affective understanding and generation abilities to produce supportive responses consistently with emotional nuances. Domain-specific adaptation and knowledge augmentation strategies significantly benefit ESC compared to zero-shot or supervised fine-tuning on a single model. MAS with scenario-immersive simulation strategy exhibits promising potential in aligning agent actions with evaluation from specialized models, which assist agents in capturing dynamic nuances and providing professional instructions or feedback in simulated scenarios. This underscores the effectiveness of collaborative intelligence and tailored knowledge enhancement for complex affective tasks that demand sophisticated emotional reasoning and generation capabilities.

\subsection{Comparative Analysis}
Structured collaboration strategies have demonstrated effectiveness across AC tasks, especially for AU tasks requiring high precision on annotated labels. Instance-level augmentation effectively addresses data scarcity, enabling models to capture fine-grained sentiment and emotion nuances. Token-level augmentation offers a computationally efficient approach to boost performance when combined with high-quality datasets or relevant knowledge. For multimodal tasks such as ERC, modal-level augmentation, which integrates signals across visual, auditory, and textual modalities, facilitates more precise emotion detection. However, effective modality alignment and fusion remain challenging. When it comes to autonomous collaboration systems, especially with advanced LLMs, they present promising yet resource-intensive solutions due to the complexity of coordination and communication. As task demands increase, designing multi-agent frameworks with domain-specific rules and specialized assistance becomes vital for robust reasoning. Accordingly, the choice of strategy hinges on the specific application context, balancing performance, resource constraints, and operational scalability.

\subsection{Future Direction}
\subsubsection{Efficient Data Augmentation}
Data scarcity remains a pervasive challenge in AC, critically impacting the ability of models to accurately capture emotional nuances present in real-world scenarios \cite{Ansari2024, Li2021LearningIS}. While LLM-based collaboration systems, notably instance-level augmentation \cite{rvisa, lai2024mtisa}, demonstrate significant promise in enhancing data diversity and quantity, further advancements are required to optimize their efficiency and governance. Future research should prioritize the development of scalable and resource-efficient data augmentation methodologies that maximize semantic richness and affective diversity while minimizing computational overhead. Incorporating domain-specific evaluators, such as scoring mechanisms or expert-in-the-loop verification, can enhance data quality and reliability, thereby facilitating real-time and scalable affective AI systems capable of continuous learning from dynamic environments.
\subsubsection{Mitigating Hallucinations}
Although LLMs exhibit remarkable understanding and generation ability in AC, they are susceptible to hallucinations \cite{Zhang2023SirensSI, Li2024ASO, lai2024mtisa} that produce plausible yet unsubstantiated or erroneous information in response. Mitigating hallucination issues in LLM outputs is essential for maintaining trustworthiness in developing robust LLM-based collaboration systems for AC, especially for AG tasks that are synthesis-driven and open-ended, requiring heavier efforts in evaluation. Techniques to improve factual grounding \cite{geng-etal-2024-survey,mahaut-etal-2024-factual} and consistency should be prioritized in affective reasoning and achieving human-like intelligence. It is promising to integrate such techniques in collaboration strategies for performance enhancement and trustworthiness in AC tasks.
\subsubsection{Multimodal Alignment}
Given that human emotional perception inherently involves multiple sensory modalities, such as visual, auditory, and textual cues, multimodal AC holds significant potential for more nuanced and accurate emotion recognition and generation. However, the complexity of aligning heterogeneous data types remains a fundamental challenge \cite{Das2023MultimodalSA, hu2024recenttrendsmultimodalaffective, GKOUMAS2021184}. Future research should focus on developing advanced multimodal fusion and alignment strategies, potentially grounded in cognitive and perceptual theories, to better integrate affective cues across modalities. Innovations in multimodal representation learning and cross-modal semantic alignment are essential for elevating the depth, precision, and contextuality of affective understanding in complex environments, thereby advancing toward more human-like emotional intelligence.
\subsubsection{Multilingual and Multicultural Understanding}
LLM-based collaboration can enhance the robustness of AC by incorporating specialized models or experts for supplementary knowledge. However, current LLMs often lack robustness in multilingual and multicultural contexts, primarily due to limited diverse datasets and culturally specific affective cues \cite{emnlp/AdilazuardaMLSA24, nips/LiC0S024, liu-etal-2024-multilingual, pawar2024surveyculturalawarenesslanguage}. Since emotional expression and interpretation are deeply influenced by language and cultural norms, extending AC models to handle multiple languages and cultural nuances effectively is imperative for fostering inclusiveness and global applicability. Future work should focus on constructing balanced, culturally diverse datasets and incorporating domain experts into annotation and refinement processes. Additionally, developing cross-cultural adaptation modules within collaborative systems can facilitate more accurate and sensitive affective interactions across diverse populations.

\subsubsection{Real-time Adaptation}
In real-world scenarios, humans perceive and express their feelings dynamically according to the shift of environments. Developing models capable of continuous, autonomous adaptation to new streams and contextual variations will foster more flexible and personalized affective interactions \cite{Bilquise2022EmotionallyIC}. Designing such systems involves overcoming significant challenges related to memory retention, contextual understanding, and stability in long-term interactions. In LLM-based collaboration systems, autonomous multi-agent systems \cite{Li2024multiagent, tran2025multiagentcollaborationmechanismssurvey, ijcai/GuoCWCPCW024} that perceive real-time feedback and adjust their responses accordingly, similar to human adaptive behavior, are promising. Integrating specialized modules, such as memory modules or tool usage modules, can enhance robustness and enable personalized, emotionally aware interactions that evolve with user needs.

\subsubsection{Ethical and Fairness Concerns}
As AC increasingly intersects with various disciplines such as psychology \cite{peiIC}, sociology \cite{vanKleef2016EditorialTS}, and education \cite{Yadegaridehkordi19}, ethical considerations become important. Data used in training and deploying affective systems often contains highly personal or sensitive information, raising concerns about privacy, bias, and fairness \cite{pieee/DevillersC23}. It is crucial to embed ethical frameworks that address bias mitigation, data privacy, and equitable treatment within the development of collaborative affective AI. Responsible innovation practices, such as transparent model auditing, bias detection, and privacy-preserving techniques, will be crucial to build trust, ensure social acceptance, and foster the ethical deployment of affective computing technologies.

%% file: src/conclusion.tex
\section{Conclusion}
\label{conclusion}

Affective computing (AC), increasingly recognized as an essential component in achieving human-like intelligence, requires both emotional and rational thinking to handle complex affective reasoning processes in real-world scenarios. The emergence of large language models (LLMs) has demonstrated remarkable abilities in understanding and generation, enabling them to function similarly to a human brain for AC tasks. However, LLMs still face cognitive limitations, such as misunderstanding cultural nuances or misinterpreting contextual emotion shifts. To address these challenges, LLM-based collaboration systems integrate specialized models with LLMs, forming teams that facilitate collaborative affective intelligence through role division, knowledge sharing, and dynamic adaptation. This survey systematically explores the emerging field of LLM-based collaborative systems in AC, highlighting their potential to emulate human-like emotional intelligence through both structured and autonomous collaboration strategies. By analyzing existing methodologies, experimental results across diverse representative affective tasks, and the inherent challenges and future directions, we underscore the transformative potential of collaborative AI in advancing affective understanding and generation. The integration of specialized models with LLMs enables more robust, adaptable, and context-aware emotional interactions, paving the way for more natural and socially intelligent human-AI interactions.